\documentclass[final]{cvpr}

\usepackage{times}

\usepackage[ruled,vlined]{algorithm2e} 
\usepackage{amsfonts,amsmath,amssymb}
\usepackage{bbm}
\usepackage{booktabs}
\usepackage{color}
\usepackage{comment}
\usepackage{dsfont}
\usepackage{enumitem}
\usepackage{epsfig}
\usepackage{graphicx}
\usepackage{makecell}
\usepackage{multirow}
\usepackage{tabularx}
\usepackage{wrapfig}
\usepackage{xcolor}
\usepackage{xspace}

\let\originalparagraph\paragraph
\renewcommand{\paragraph}[2][.]{\vspace{-\baselineskip}\originalparagraph{#2#1}}

\newcolumntype{L}{>{\raggedright\arraybackslash}X}
\newcolumntype{C}{>{\centering\arraybackslash}X}

\makeatletter
\DeclareRobustCommand\onedot{\futurelet\@let@token\@onedot}
\def\@onedot{\ifx\@let@token.\else.\null\fi\xspace}
\def\eg{\emph{e.g}\onedot} 
\def\ie{\emph{i.e}\onedot}

\def\etal{\emph{et al}\onedot}
\makeatother

\definecolor{mkcolor}{RGB}{255,0,255}

\definecolor{vovacolor}{RGB}{255,0,0}

\definecolor{mhcolor}{RGB}{0,128,0}

\definecolor{nmcolor}{RGB}{0,0,255}

\definecolor{sidcolor}{RGB}{128,0,128}

\definecolor{leocolor}{RGB}{255,128,0}

\graphicspath{{figures/}}

\usepackage[pagebackref=true,breaklinks=true,colorlinks,bookmarks=false]{hyperref}

\def\cvprPaperID{1247} 
\def\confYear{CVPR 2021}

\begin{document}

\title{Joint Learning of 3D Shape Retrieval and Deformation}

\author{Mikaela Angelina Uy$^1$~~~Vladimir G. Kim$^2$~~~Minhyuk Sung$^3$~~~Noam Aigerman$^2$
	\vspace{0.1cm}\\
	Siddhartha Chaudhuri$^{2,4}$~~~Leonidas Guibas$^{1}$
	\vspace{0.2cm}\\
	$^1$Stanford University~~~$^2$Adobe Research~~~$^3$KAIST~~~$^4$IIT Bombay
	\vspace{-0.2cm}\\
}

\maketitle
\vspace{-3pt}

\begin{abstract}
\vspace{-7pt}
We propose a novel technique for producing high-quality 3D models that match a given target object image or scan. Our method is based on retrieving an existing shape from a database of 3D models and then deforming its parts to match the target shape. Unlike previous approaches that independently focus on either shape retrieval or deformation, we propose a joint learning procedure that simultaneously trains the neural deformation module along with the embedding space used by the retrieval module. This enables our network to learn a deformation-aware embedding space, so that retrieved models are more amenable to match the target after an appropriate deformation. In fact, we use the embedding space to guide the shape pairs used to train the deformation module, so that it invests its capacity in learning deformations between meaningful shape pairs. Furthermore, our novel part-aware deformation module can work with inconsistent and diverse part-structures on the source shapes. We demonstrate the benefits of our joint training not only on our novel framework, but also on other state-of-the-art neural deformation modules proposed in recent years. Lastly, we also show that our jointly-trained method outperforms various non-joint baselines. 
\end{abstract}
\vspace{-15pt}

\section{Introduction}
\vspace{-3pt}
Creating high-quality 3D models from a reference image or a scan is a laborious task, requiring significant expertise in 3D sculpting, meshing, and UV layout. While neural generative techniques for 3D shape synthesis hold promise for the future, they still lack the ability to create 3D models that rival the fidelity, level of detail, and overall quality of artist-generated meshes~\cite{what3d_cvpr19}. Several recent techniques propose to directly retrieve a high-quality 3D model from a database and deform it to match a target image or point cloud, thereby approximating the target shape while preserving the quality of the original source model. These prior methods largely focus on one of two complementary subproblems: either retrieving an appropriate mesh from a database~\cite{yanyang, dahnert2019embedding}, or training a neural network to deform a source to a target~\cite{CycleConsistency,wang20193dn,Yifan:NeuralCage:2020,Sung:2020}. In most cases, the static database mesh most closely matching the target is retrieved, and then deformed for a better fit~\cite{ishimtsev2020caddeform}. The retrieval step is not influenced by the subsequent deformation procedure, and thus ignores the possibility that a database shape with different global geometry nevertheless possess local details that will produce the best match {\em after} deformation.

\begin{figure}[t]
    \centering
    \includegraphics[width=\linewidth]{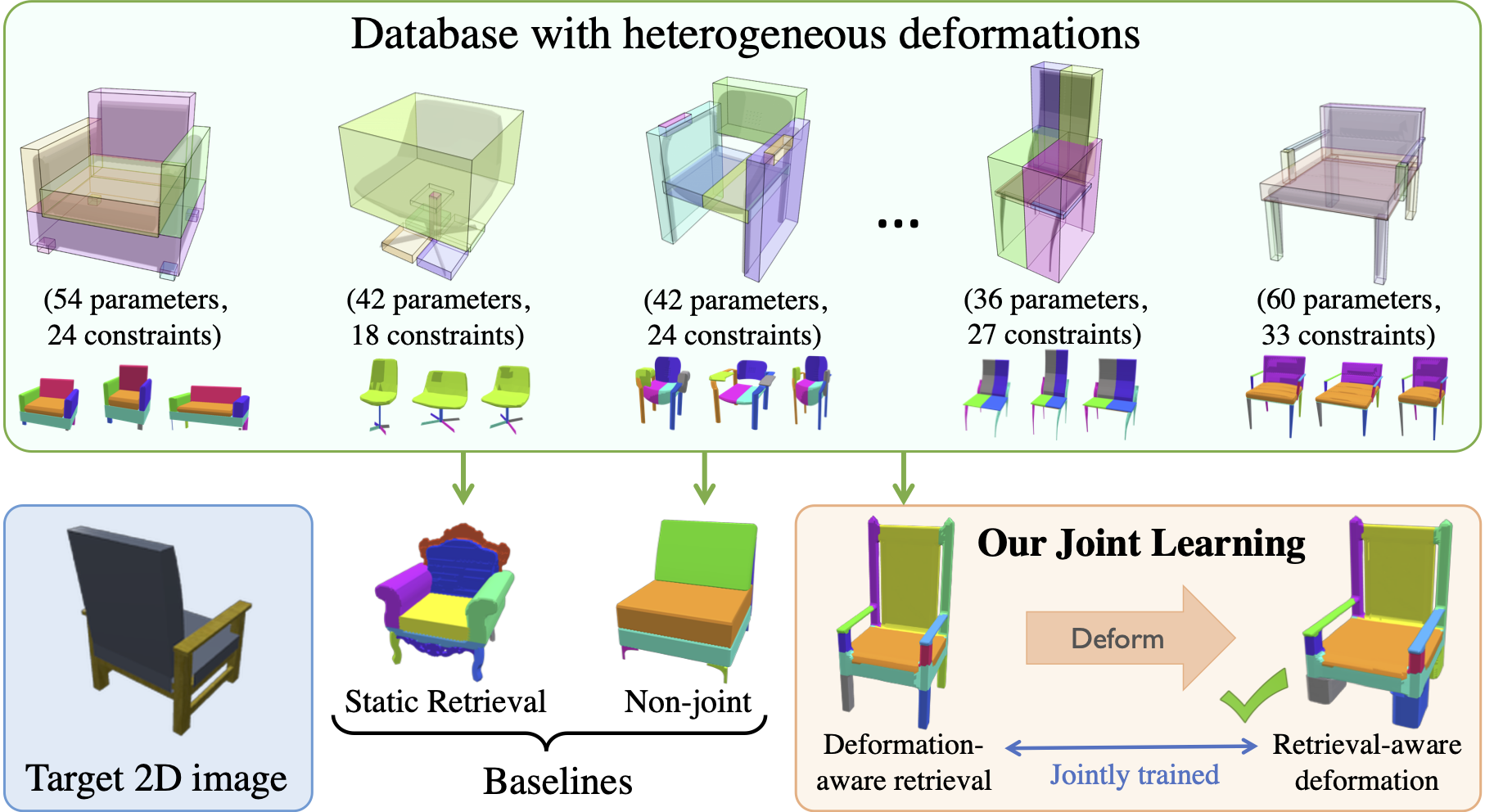}
    \vspace{-.5cm}
    \caption{Given an input target we use jointly-learned retrieval and deformation modules to find a source model in a heterogeneous database and align it to the target. We demonstrate that our joint learning outperforms static retrieval and non-joint baselines.}
    \vspace{-.5cm}
    \label{fig:teaser}
\end{figure}

Only a few works explicitly consider {\em deformation-aware} retrieval~\cite{schulz2017_param,uy-deformawareretrieval-eccv20}. However, in these works the deformation module is a fixed, non-trainable black box, which requires complete shapes (and not e.g., natural images or partial scans) as targets, does not handle varying shape structures across the database, may necessitate time-consuming, manually-specified optimization of a fitting energy, exhaustive enumeration of deformed variants, and does not support back-propagating gradients through it for directly translating deformation error to retrieval error.

In this paper, we argue that retrieval and deformation should be {\em equal citizens in a joint problem}. Given a database of source models equipped with some parametric representation of deformations, our goal is to learn how to retrieve a shape from the database and predict the optimal deformation parameters so it best matches a given target. A key feature of our method is that both retrieval and deformation are {\em learnable} modules, each influencing the other and trained jointly. While the benefit of deformation-aware retrieval has been explored previously, we contribute the notion of {\em retrieval-aware deformation}: our learnable deformation module is optimized for fitting retrieved shapes to target shapes, and does not try to be a general-purpose algorithm for arbitrary source-target pairs. Thus, the retrieval module is optimized to retrieve sources that the deformation module can fit well to the input target, and the deformation module is trained on sources the retrieval module predicts for the input target, thereby letting it optimize capacity and learn only meaningful deformations.

The robustness of the joint training enables us to devise a more elaborate deformation space. Specifically, we devise a {\em differentiable, part-aware deformation function} that deforms individual parts of a model while respecting the part-to-part connectivity of the original structure (Figure~\ref{fig:teaser}). Importantly, it accommodates varying numbers of parts and structural relationships across the database, and does not require part labels or consistent segmentations. It can work with automatically-segmented meshes and even multiple differently segmented instances of the same source shape. We propose a way to encode each part in each source and to enable a general MLP to predict its deformation regardless of the part count. This holistic view of joint retrieval and deformation is especially important when considering heterogeneous collections of shapes ``in the wild'' that often vary in their part structure, topology, and geometry. These require different deformation spaces for different source models, which the retrieval module must be aware of.

We evaluate our method by matching 2D image and 3D point cloud targets. We demonstrate that it outperforms various baselines, such as vanilla retrieval~\cite{yanyang}, or deformation-aware retrieval using direct optimization for deformation~\cite{uy-deformawareretrieval-eccv20}, or a fixed, pre-trained, neural deformation module (i.e. omitting joint training). We also show that our method can be used even with imperfect and inconsistent segmentations produced automatically. Finally, we show that even with a different deformation module (e.g., Neural Cages~\cite{Yifan:NeuralCage:2020}), our joint training leads to better results.

\section{Related Works}
\vspace{\baselineskip}

\paragraph{Deep Learning for Shape Generation}
Many neural techniques have been proposed recently for learning generative latent representations for 3D shapes, modeling geometry as  implicit functions~\cite{deepsdf, Occupancy_Networks, chen2018implicit_decoder}, atlases~\cite{atlasnet}, volumetric grids~\cite{cnncomplete, jiajun3dgan}, point clouds~\cite{Achlioptas:2018, pointflow, pointsetgen}, and meshes~\cite{wang2018pixel2mesh, wen2019pixel2mesh++}. These models tend to under-perform on topologically complex objects with intricate part structures. Thus, other techniques focus on factorized representation, where variations in structure are modeled separately from the geometry~\cite{grass, gaosdmnet2019, mo2019structurenet}. These generative techniques are commonly used jointly with 2D CNNs~\cite{qi2016volumetric} or shape encoders~\cite{Zhirong15CVPR} to enable creating a shape based on some partial observations, such as a natural image~\cite{meshrcnn} or a point scan~\cite{Dai2019Scan2MeshFU}.  A simple shape retrieval~\cite{yanyang} could also be viewed as the simplest version of such a shape generator, where the system simply returns the nearest neighbor in the latent space, in fact, offering a strong baseline to other generative techniques~\cite{what3d_cvpr19}. 

\paragraph{Deformation-Aware Retrieval}
Direct retrieval has the advantages of producing stock-quality meshes~\cite{Warehouse, TurboSquid}, however, unless the database contains all possible objects, might not produce a good fit for an encoded target. Prior works~\cite{Nan2012ASA, schulz2017_param, uy-deformawareretrieval-eccv20} address this issue by additionally deforming, \ie fitting, the retrieved shape to the desired target. One approach is to exhaustively deform all shapes in the database to the target and select the best fit~\cite{Nan2012ASA}, but is however computationally expensive. Schulz \etal~\cite{schulz2017_param} alleviates this by retrieving parametric models by representing each as a set of points and bounded tangent planes, thus enabling retrieval before the fitting process. Leveraging on deep networks, Uy \etal~\cite{uy-deformawareretrieval-eccv20} use a deep embedding to retrieve a shape and then separately deform it to the target by directly optimizing the ARAP~\cite{Igarashi:2005} loss. Their method is limited to full shapes as targets as direct optimization is not possible with partial scans~\cite{Avetisyan:2019:Scan2CAD} or natural images. They further observe that the retrieval network needs to be aware of the deformation step to retrieve a more appropriate source. We extend their approach in several ways. First, we demonstrate that one can use retrieve-and-deform method with a neural deformation technique, allowing us to handle natural images as inputs. Second, we propose a novel joint training process, which enables us to train our deformation module to be more suitable for the kind of pairs of shapes that are being retrieved. And third, we propose a novel neural deformation module that is especially suitable for heterogeneous shape collections with topological and structural variations. 


\paragraph{3D Deformation} 
Deforming a source 3D model to a target is one of the fundamental problems in geometry processing. If target is a full shape, direct optimization techniques can be employed~\cite{HuangNR-ICP, Sorkine:2007, mvc}, as well as human-made~\cite{Kim13, GanapathiSubramanian2018ParsingGU, Xu13z.:style-content, zheng} shapes. One can only directly optimize if a target is a full shape, however if it of a different modality, such as image or partial scan, one needs to employ priors~\cite{alignet}. Neural techniques have been used to learn such deformation priors from collections of shapes, representing deformations as volumetric warps~\cite{Jack:2018, deformnet, Yumer:2016}, cage deformations~\cite{Yifan:NeuralCage:2020}, vertex-based offsets~\cite{wang20193dn, CycleConsistency} or flow-based approaches~\cite{jiang2020shapeflow}. To make learning easier, these techniques typically assume homogeneity in the sources and represent the deformation with the same number of parameters for each source, \ie grid control points~\cite{Jack:2018}, cage mesh~\cite{Yifan:NeuralCage:2020} or number of vertices~\cite{wang20193dn}. These assumptions make them less suitable for heterogeneous databases of sources with significant structural variations at the part level. We extend the part-level reasoning that proved to be effective for other problems~\cite{Mo_2019_CVPR, xu_sig11, 3dsurveySTAR} to neural deformation, by proposing a novel module that can learn source-specific deformations, and handle cases when sources can have different number of deformation parameters to account for part variability.

\begin{figure*}[t]
    \centering
    \includegraphics[width=\linewidth]{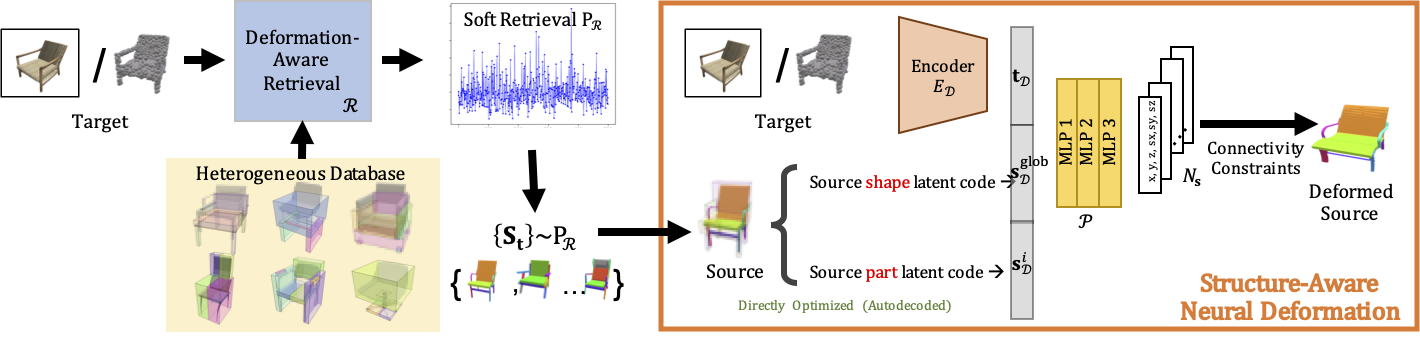}
    \caption{During training, given a target image or a point cloud and a database of deformable sources, we retrieve a subset of source models based on their proximity in the retrieval space, and use the structure-aware deformation module (right) to fit each source. Our deformation module uses encoded target, global and per-part source codes to predict per-part deformation parameters.
    }
    \label{fig:network}
    \vspace{-\baselineskip}
\end{figure*}

\section{Method}
\vspace{\baselineskip}

\paragraph{Overview} We assume to possess a database of parametric \emph{source} models $\mathbf{s}\in\mathbf{S}$, and we aim to jointly train a deformation and retrieval module to choose a source and deform it to fit a given target $\mathbf{t}$ (an image or a point cloud), with respect to a fitting metric $\mathcal{L}_\text{fit}$ (we use chamfer in all experiments). Each source also has parameters defining its individual deformation space, that are optimized during training. 

Our deformation module is designed to enable a different deformation function $\mathcal{D}_\mathbf{s}$ for each source $\mathbf{s}$, based on its parts. The retrieval module uses embeddings of the sources and the target into a latent space $\mathcal{R}$, where it retrieves based on a distance measure $\mathrm{d}_\mathcal{R}$, which enables the retrieval of the source shape that best fits to the target \emph{after} deformation.

The training consists of optimizing the latent retrieval space $\mathcal{R}$ and the deformation functions $\{\mathcal{D}_\mathbf{s}\}$:
$$
\vspace{-3pt}
\min \mathcal{L}_\text{fit} \left( \mathcal{D}_{s'}(\mathbf{t}),  \mathbf{t}_\text{true}\right),
$$
where $s'$ is the closest source to target $\mathbf{t}$ in latent space, w.r.t the distance measure $d_{\mathcal{R}}(\mathbf{s'}, \mathbf{t})$, and $\mathbf{t}_\text{true}$ is the corresponding true shape.


We progress by first explaining in Section \ref{sec:joint} how we design our framework and optimization in a way that is differentiable and enables the deformation and retrieval modules to propagate information from one to the other. We then detail our novel deformation module and how it enables source-specific deformations in Section \ref{sec:deform}, and conclude by describing the retrieval module in Section \ref{sec:retrievald}.

\subsection{Joint Deformation and Retrieval Training}
\label{sec:joint}
\vspace{-3pt}
It is critical for our approach to optimize the parameters of $\mathcal{R}$ and $\{\mathcal{D}_\mathbf{s}\}$ jointly. First, it enables the deformation module of each source to efficiently utilize its capacity and specialize on relevant targets that it could fit to. Second, it allows the retrieval module to create a deformation-aware latent space where sources are embedded closer to the targets they can deform to. 

\paragraph{Soft Retrieval for Training}

The retrieval module embeds the sources and the target in the latent retrieval space $\mathcal{R}$. 
The proximity in latent space is used to define a biased distribution that can be loosely interpreted as the probability of source $\mathbf{s}$ being deformable to $\mathbf{t}$:
\vspace{-3pt}
\begin{equation} 
\mathrm{P}_\mathcal{R}(\mathbf{s}, \mathbf{t})=\mathrm{p}(\mathbf{s} ; \mathbf{t}, \mathbf{S}, \mathrm{d}_\mathcal{R}, \mathrm{\sigma}_0),
\label{eq:soft_retrieval}
\end{equation}
where
\vspace{-3pt}
\begin{equation*}
\mathrm{p}(\mathbf{s} ; \mathbf{t}, \tilde{\mathbf{S}}, \tilde{\mathrm{d}}, \tilde{\mathrm{\sigma}}) = 
\frac{\exp(-\tilde{\mathrm{d}}^2(\mathbf{s}, \mathbf{t})/\tilde{\mathrm{\sigma}}^2(\mathbf{s}))}{\sum_{\mathbf{s}' \in \tilde{\mathbf{S}}} \exp(-\tilde{\mathrm{d}}^2(\mathbf{s}', t)/\tilde{\mathrm{\sigma}}^2(\mathbf{s})) },
\end{equation*}
$\tilde{\mathrm{d}}: \left( \mathbf{S} \times \mathbf{T} \right) \rightarrow \mathbb{R}$ is a distance function between a source and a target ($\mathbf{T}$ is the target space), and $\tilde{\mathrm{\sigma}}: \mathbf{S} \rightarrow \mathbb{R}$ is a potentially source-dependent scalar function. Though, $\sigma_0 (\cdot)=100$ is a constant set for all experiments.

Instead of choosing the highest-scoring source according to the probability  $\mathrm{P}_\mathcal{R}$,  we perform soft retrieval and \emph{sample} $K=10$ retrieval candidate sources from the distribution:
\vspace{-3pt}
\begin{equation*}
\mathbf{s}_i \sim \mathrm{P}_\mathcal{R}(\mathbf{s}, \mathbf{t}), \forall i \in \{1,2,...,K\}.
\label{eq:prob}
\end{equation*}

The candidates $\mathbf{S}_\mathbf{t}=\{\mathbf{s}_1,...,\mathbf{s}_K\}$ sampled via our soft retrieval are then used to train both our retrieval module to learn $\mathcal{R}$ and deformation module for source-depedent deformation functions $\{\mathcal{D}_\mathbf{s}\}$.

The soft retrieval is crucial for our training: 1) adding randomness to the retrieval ensures that the latent space is optimized with respect to both high-probability instances and low-probability ones, that may reverse roles as the deformation module improves. 2) On the other hand, training the deformation module with a bias towards high-probability sources and not random ones ensures it is aware of the retrieval module and expands its capacity on meaningful matches.

\paragraph{Training} We train the two modules jointly in an alternating fashion, keeping one module fixed when optimizing the other, and vice versa, in successive iterations. To train the retrieval module, we deform the candidate sources and compute their fitting losses to the target. We update our latent space $\mathcal{R}$ by penalizing the discrepancy between the distances in the retrieval space $d_\mathcal{R}$ and the post-deformation fitting losses $\mathcal{L}_\text{fit}$ using softer probability measures estimated from the distances of the sampled candidates:
\vspace{-3pt}
\begin{equation}
\mathcal{L}_\text{emb} = \sum_{k=1}^K |\
\mathbf{p}(\mathbf{s}_k, \mathbf{t}, \mathbf{S}_\mathbf{t}, d_\mathcal{R}, \sigma_0)
- \mathbf{p}(\mathbf{s}_k, \mathbf{t}, \mathbf{S}_\mathbf{t}, d_\text{fit}, \sigma_k)|,
\label{eq:loss_embed}
\end{equation}
where
\vspace{-3pt}
\begin{equation}
d_\text{fit}(\mathbf{s}, \mathbf{t})=\mathcal{L}_\text{fit}(\mathcal{D}_\mathbf{s}(\mathbf{t}), \mathbf{t}_\text{true}),
\end{equation}
and $\sigma_k$ is a source-dependent scalar representing the predicted range of variations of each source model $\mathbf{s}\in\mathbf{S}$, which is also learned.
For the deformation module, we update the deformation functions $\{\mathcal{D}_{\mathbf{s}_k}\}$ for the $K$ biased samples by minimizing the post-deformation fitting losses weighted by their soft probability measures:
\vspace{-3pt}
\begin{equation}
\mathcal{L}_\text{def} = \sum_{k=1}^K
\mathbf{p}(\mathbf{s}_k, \mathbf{t}, \mathbf{S}_\mathbf{t}, d_\mathcal{R}, \sigma_0)
\mathcal{L}_\text{fit} (\mathcal{D}_{\mathbf{s}_k}(\mathbf{t}), \mathbf{t_\text{true}}).
\label{eq:loss_embed}
\end{equation}

This weighting scheme puts greater weight on sources that are closer to the target in the embedding space, thus further making the deformation module aware of the retrieval module, and allowing it to specialize on more amenable sources with respect to the training target.

\paragraph{Inner Deformation Optimization}
To enforce the deformation module to perform more significant deformations, at each training iteration we use the deformation network's current output for the given source and target that consists of parameters for the deformation, and directly run SGD on the deformation parameters until convergence of the fitting loss. We then measure the least-square error between the deformation network's output and the optimized parameters, and train the module by backpropagating this error, hence enabling the network to learn stronger deformations and getting a better estimate for how well the source could be aligned to the target after the deformation. See the supplementary for the full details.

\subsection{Structure-Aware Neural Deformation}
\label{sec:deform}
\vspace{-3pt}

While our joint training approach described in Section~\ref{sec:joint} is generic and can work well with different parameterization of deformations, its greatest advantage is that it enables our deformation space to vary greatly between each source without having the deformation module learn subpar deformations.   We thus devise a deformation module with a heterogeneous space of part-based deformations as shown in Figure~\ref{fig:teaser}, which vary per each source, a necessary feature if one wants to tailor the deformations to be restricted to preserve and adjust part structures.

To get meaningful parts, we use manual segmentations from PartNet~\cite{Mo_2019_CVPR} or automatic segmentations (preprocessing) of ComplementMe~\cite{Sung:2017}, produced by grouping connected components in raw meshes. Our deformation module predicts a simple deformation consisting of translation and axis-aligned scaling for each part in a source model. See supplementary for the details on the prediction. The number of parts for different sources vary, making the deformation functions source-dependent $\{\mathcal{D}_\mathbf{s}\}$. We abuse the notation a bit and let $\mathcal{D}$ denote our deformation module.

We propose to use a neural network which can be applied to each part separately, thus making it applicable to models with varying part-constellations, as opposed to previous methods. Namely, we assign to each source a global code $\mathbf{s}_\mathcal{D}^\text{glob} \in \mathbb{R}^{n_1}$, and for each part within the shape, we assign a local code $\mathbf{s}_\mathcal{D}^{i=1...N_s}\in \mathbb{R}^{n_2}$.   The target is encoded via an encoder (PointNet~\cite{qi2016pointnet} for point clouds and ResNet~\cite{He2016DeepRL} for images) into a latent vector $\mathbf{t}_\mathcal{D} = E_\mathcal{D}(\mathbf{t}) \in \mathbb{R}^{n_3}$. We set $n_1=n_3=256$ and $n_2=32$ for all experiments. The global, local, and target codes are concatenated and fed to a lightweight 3-layer MLP (512, 256, 6), $\mathcal{P}$, which outputs the deformation parameters of the corresponding part. The deformation parameters of all parts are then used to obtain the final deformed source shape. Each source's global and local codes are optimized in an auto-decoder fashion during the training of the deformation module. Figure~\ref{fig:network} (right) illustrates our module. We additionally add a symmetry loss in training our deformation module to enforce bilateral symmetry of the output deformed shapes as regularization, more details are found in the supplementary. \\

\paragraph{Connectivity Constraints}
We further take advantage of our joint-training's robustness to heterogeneous deformation spaces, and add part-connectivity constraints. We achieve this by introducing a layer that receives a deformation and projects it onto the space of contact-preserving deformations, via a simple linear transformation. Contacts are defined between pairs of connected parts where each pair introduces a set of constraints. The source models have different sets of connected parts, and hence a different number and set of constraints, as illustrated in Figure~\ref{fig:teaser}, making the deformation functions $\{\mathcal{D}_\mathbf{s}\}$ even more source-dependent. More details are found in the supplementary.

\subsection{Retrieval in Latent Space}
\label{sec:retrievald}
\vspace{-3pt}
The retrieval space $\mathcal{R}$ is defined similarly to Uy et al.~\cite{uy-deformawareretrieval-eccv20}, and we provide relevant technical details in this section for completeness. We use a PointNet or ResNet encoder to get the latent code of the target: $\mathbf{t}_\mathcal{R} = E_\mathcal{R}(\mathbf{t}) \in \mathbb{R}^{n_4}$ with $n_4=256$. The sources are represented as regions in the latent space, defined by a center code $\mathbf{s}_\mathcal{R}\in \mathbb{R}^{n_4}$ and a variance matrix $\mathbf{s}_\mathcal{R}^v \in \mathbb{R}^{n_4 \times n_4}$ that defines the egocentric distance field. The variance matrix is diagonal positive definite, with the positivity enforced by the sigmoid activation function. We define the distances in the retrieval space as:
\begin{equation}
d(\mathbf{s}, \mathbf{t})_\mathcal{R} = \sqrt{(\mathbf{s}_{\mathcal{R}}-\mathbf{t}_{\mathcal{R}})^{T}\mathbf{s}_{\mathcal{R}}^v(\mathbf{s}_{\mathcal{R}}-\mathbf{t}_{\mathcal{R}})}.
\end{equation}

During training we optimize the parameters of the encoder $E_\mathcal{R}(\mathbf{t})$ as well as latent codes and variances for each source, $\mathbf{s}_\mathcal{R},\mathbf{s}_\mathcal{R}^v$. $\mathbf{s}_\mathcal{R}$ is obtained by feeding the default shape of source model $\mathbf{s}$ to encoder $E_\mathcal{R}(\mathbf{t})$. Different from Uy et al.~\cite{uy-deformawareretrieval-eccv20}, we optimize $\mathbf{s}_\mathcal{R}^v$ in an auto-decoder fashion, since we want to represent the deformation space of the source rather than 
its geometry. This allows us to handle sources with similar geometry but different parameterizations.
\vspace{-2.5pt}
\section{Results}
\vspace{-3pt}
In this section we discuss our data sources and evaluation metric and provide thorough experiments with image (Sec~\ref{sec:image-to-mesh}) and point cloud (Sec~\ref{sec:points-to-mesh}) targets. 
\vspace{-1pt}

\paragraph{Datasets and Evaluation Metric}
We evaluate our method on the three furniture categories in the ShapeNet dataset~\cite{ShapeNet} chairs (6531), tables (7939) and cabinets (1278). For our database of deformable source models, we use manually- and automatically-segmented shapes from two different datasets. Manually-segmented shapes come from the finest level of PartNet hierarchy~\cite{Mo_2019_CVPR}, and we select random 10\% of the data as our sources. Automatically-segmented shapes come from two pre-analyzed classes in ComplementMe~\cite{Sung:2017} (chairs and tables), and we pick 200 random models for each. We remove the selected sources from the database, and use remaining models as training (80\%) and testing (20\%) targets. To demonstrate the practical utility of our method, we also test our trained networks on product images and 3D scans. 

We represent the shapes by uniformly sampling 2048 points. For the image experiments, we render 24 uniformly-sampled viewpoints, and pick a random view at each iteration during training. In all cases our true targets and deformed sources are represented as point clouds, and points-to-points distances are used for training and evaluation.

\begin{figure}[t]
    \centering
    \includegraphics[width=\linewidth]{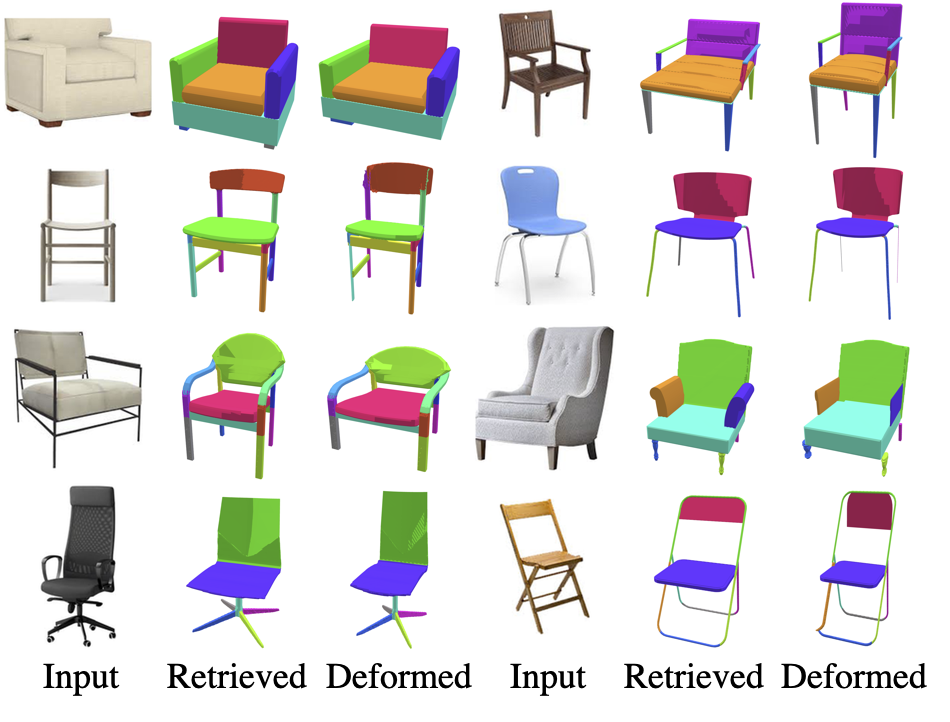}
     \vspace{-7mm} 
    \caption{We test our trained method on online product images.}
    \label{fig:naturalimages}
    \vspace{-\baselineskip}
\end{figure}

\subsection{Image-to-Mesh}
\label{sec:image-to-mesh}
\begin{figure*}[t]
    \centering
    \includegraphics[width=0.95\linewidth]{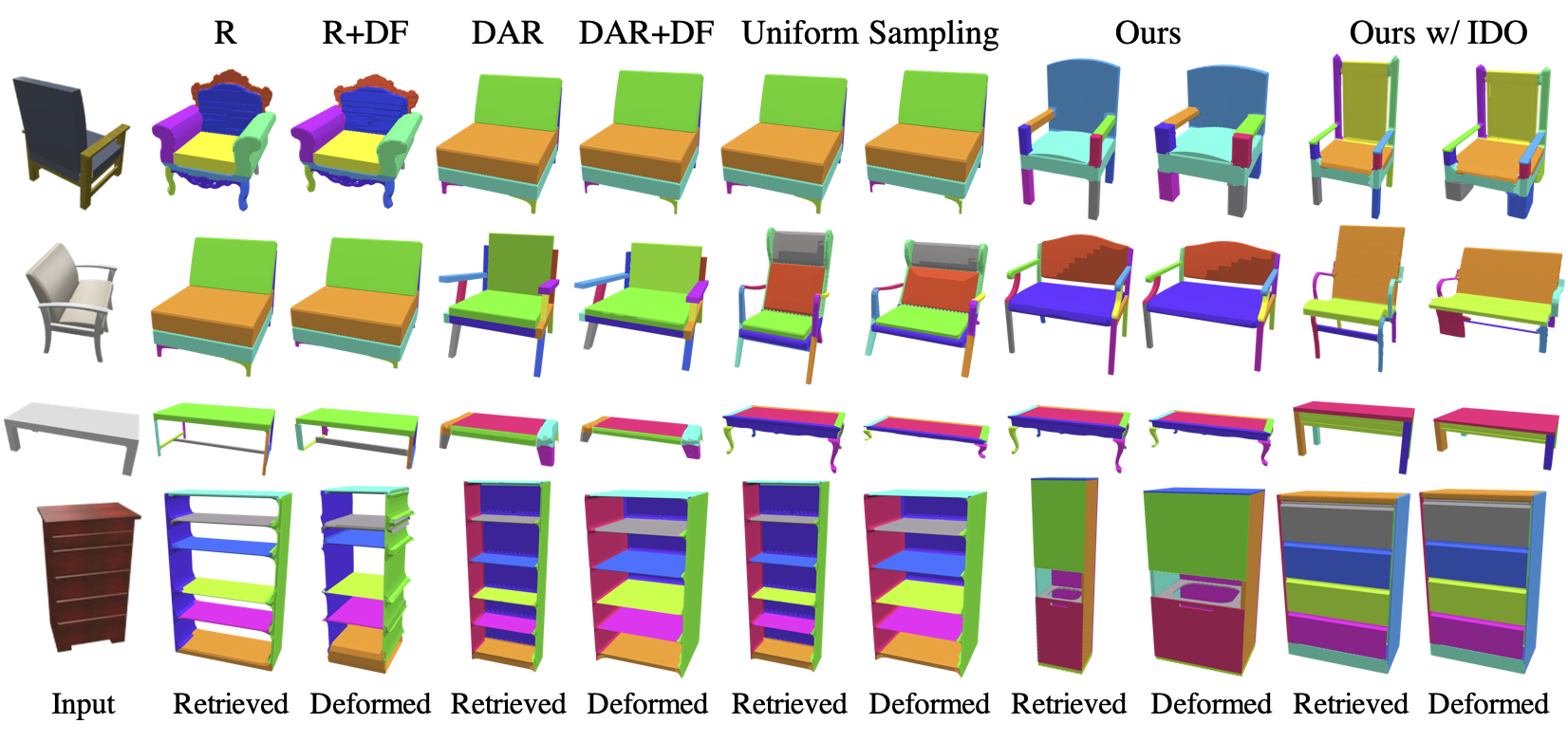}
    \vspace{-4mm} 
    \caption{Comparison between our approach and baselines for the image-to-mesh experiment. 
    }
    \label{fig:images}
    \vspace{-\baselineskip}
\end{figure*}

\vspace{-2pt}
We first test our system on product images ``in the wild'' as well as images from our test set and show qualitative results for retrieval and deformation in Figures~\ref{fig:naturalimages} and~\ref{fig:images}. Note how retrieved results have compatible structure to the input, which then enables the deformation technique to match the source to the target.
We quantitatively evaluate performance of our method and report chamfer distances in Table~\ref{tbl:image2cad} (\textbf{Ours}) together with the chamfer distances with the inner deformation optimization (\textbf{Ours w/ IDO}). Since IDO step described significantly increases training time, we do not use it in ablations and comparisons.\\
\vspace{-2.5pt}

\begin{table}
{
\footnotesize
\centering
\setlength\tabcolsep{2pt}
\begin{tabularx}{\linewidth}{c|C|C|C}
    \toprule
    &Chair&Table&Cabinet\\
    \midrule
    R & 1.926 & 2.235 & 2.228 \\
    R+DF & 1.969 & 2.705 & 2.035 \\
    \midrule
    DAR (\scriptsize{Retrieval Only}) & 1.345 & 2.058& 3.489 \\
    DAR+DF & 1.216 & 1.621 & 1.333\\
    \midrule
    Uniform Sampling & 1.118 & 1.486& 1.318\\
    \textbf{Ours} & \textbf{1.005} & \textbf{0.970} & \textbf{1.220} \\
    \textbf{Ours w/ IDO} & \textbf{0.976} & \textbf{0.935} & \textbf{1.141} \\
    \bottomrule
\end{tabularx}
\caption{Comparing our method to various baselines and ablations on image-to-mesh benchmark (chamfer distances, $\times 10^{-2}$).
}
\label{tbl:image2cad}
}
\vspace{-\baselineskip} 
\end{table}

\paragraph{Retrieval Baselines}
We compare our method to a vanilla image-to-shape retrieval technique~\cite{yanyang} (denoted by \textbf{R}). This baseline first constructs the latent space by projecting shape-to-shape chamfer distance matrix to 256-dimensional space via MDS, and then trains a ResNet~\cite{He2016DeepRL} encoder to map images to that latent space with $L2$-loss. 
Since any retrieval baseline can also work with a pre-trained neural deformation, we also train our structure-aware deformation module on random pairs of shapes (i.e., ablating the joint training procedure) and report results with neural deformation applied to the retrieved results (\textbf{R+DF}). 
Since this vanilla baseline retrieves only based on geometric similarity and does not account for deformation, the retrieved shapes may not deform to targets well. Hence, there is no improvement when deforming with the pre-trained deformation function.

The second retrieval baseline is the deformation-aware retrieval~\cite{uy-deformawareretrieval-eccv20}, where we also use our structure-aware deformation module pre-trained on random pairs. For this baseline we report results for retrieval (\textbf{DAR}) as well as deformation (\textbf{DAR+DF}). 
Our results show that being 
deformation-aware is not sufficient, and it is important for deformation module to be trained with retrieved shapes. \\
\vspace{-2.5pt}
\paragraph{Biased Sampling Ablation}
Our joint training benefits from biasing sampling of retrieval targets (Eq.~\ref{eq:soft_retrieval}). To ablate this, we sample from a uniform distribution, i.e., each source is sampled with equal probability during training. In this setting, while the retrieval and deformation modules are still trained together,
they are less aware of which samples are most relevant at inference time and thus yield higher errors (see \textbf{Uniform Sampling} in Table~\ref{tbl:image2cad}). 

\paragraph{Improvement in Deformation Module}
In addition to holistic improvement to the final output, we would like to evaluate the effect of joint training on deformation module. To do this, we use \emph{oracle retrieval} where for each test target, we deform all sources and pick the one with the smallest fitting error. Our joint training allows the deformation module to specialize on targets that are a good fit. Thus, as shown in Table~\ref{tbl:deform_image}, our method achieves the lowest fitting error for the best-fit sources with respect to the deformation module trained on all pairs (\textbf{DF}), and the deformation module trained without the biased sampling (\textbf{Uniform Sampling}).


\begin{table}
\footnotesize
\centering
\setlength\tabcolsep{2pt}
\begin{tabularx}{\linewidth}{c|C|C|C}
    \toprule
    &Chair&Table&Cabinet\\
    \midrule
    DF & 0.748 & 0.702 & 0.706 \\
    Uniform Sampling & 0.755 & 0.690 & 0.701\\
    \textbf{Ours} & \textbf{0.681} & \textbf{0.584} & \textbf{0.675}\\
    \textbf{Ours w/ IDO} & \textbf{0.669} & \textbf{0.533}  &\textbf{ 0.689} \\
    \bottomrule
\end{tabularx}
\caption{Improvement in deformation module for image-to-mesh task with oracle retrieval due to joint training (chamfer $\times 10^{-2}$).}
\label{tbl:deform_image}
\vspace{-1.2\baselineskip} 
\end{table}


\subsection{Points-to-Mesh}
\label{sec:points-to-mesh}
We also test our method on point cloud targets. We first show qualitative results with \emph{real} noisy and partial 3D scans in Scan2CAD dataset~\cite{Avetisyan:2019:Scan2CAD}.
Figure~\ref{fig:scan2cad} show some examples, and more are in the supplementary.
As shown, given an incomplete scan, with missing parts and a noise, our approach still correctly retrieves and deforms a source database model to output a clean and complete mesh to match the scan. Our structure-aware neural deformation leverages learned shape priors to complete missing regions.


\begin{figure}[t]
    \centering
    \includegraphics[width=\linewidth]{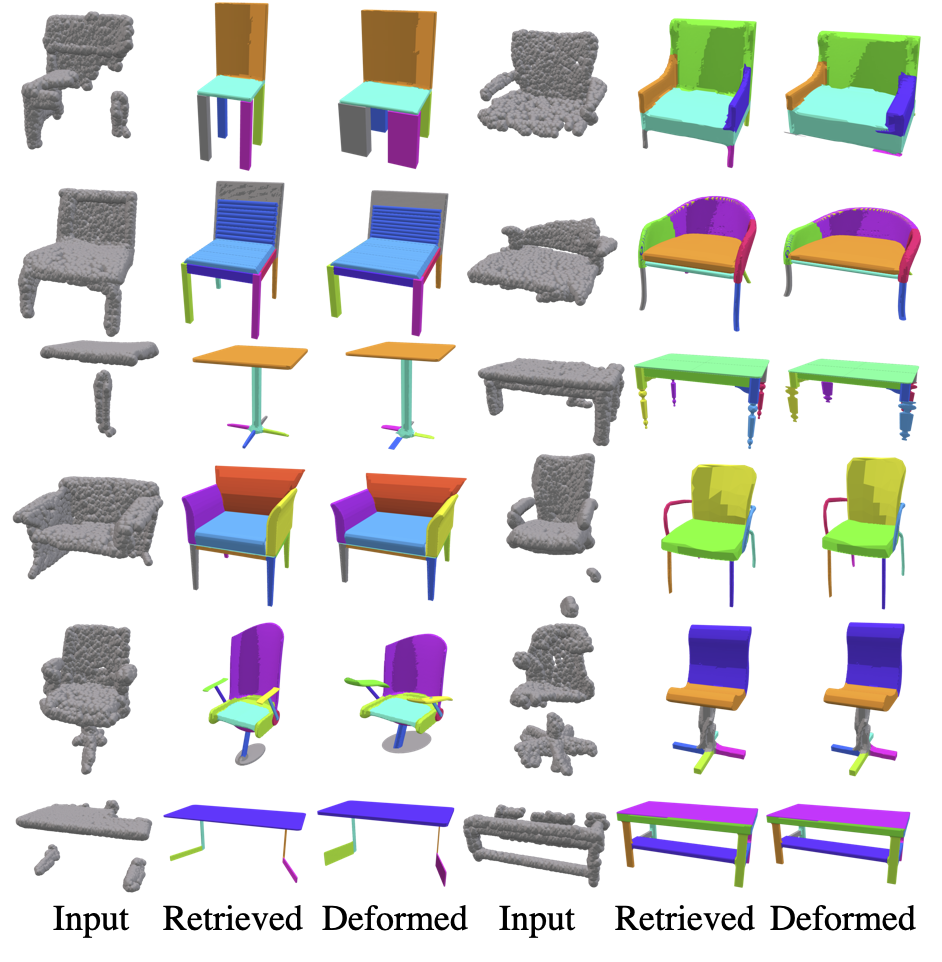}
    \vspace{-6mm} 
    \caption{Our approach tested on real scans from the Scan2CAD dataset \cite{Avetisyan:2019:Scan2CAD}. 
    }
    \label{fig:scan2cad}
    \vspace{-\baselineskip}
\end{figure}

\begin{figure}[t]
    \centering
    \includegraphics[width=\linewidth]{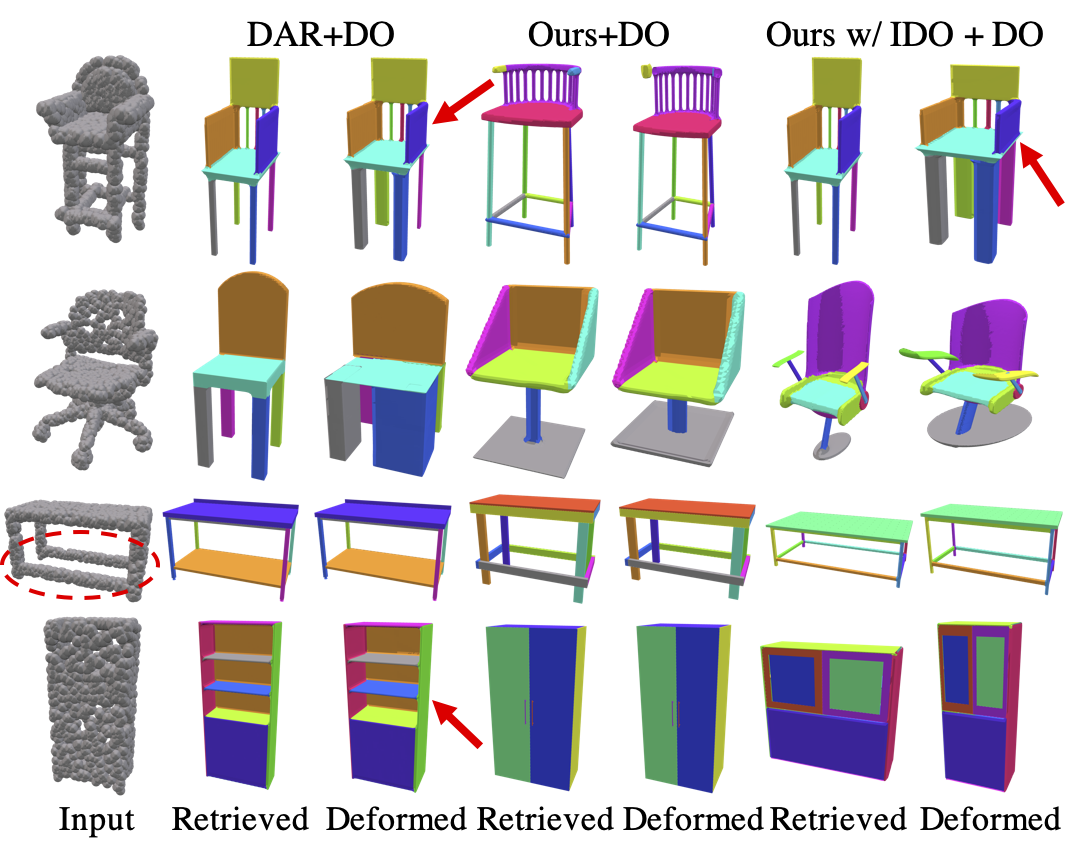}
    \vspace{-6mm} 
    \caption{Comparison between our approach and baselines for the point-cloud-to-mesh experiment. 
    }
    \label{fig:pcicp}
\end{figure}

We also provide qualitative and quantitative results on our test set of point clouds sampled from ShapeNet meshes in Table~\ref{tbl:pointcloud2cad} and Figure~\ref{fig:pcicp}. As in the previous section, we report our results (\textbf{Ours}) along with our method with the inner direct optimization step (\textbf{Ours w/ IDO}). Since our input are point clouds, similar to prior work~\cite{uy-deformawareretrieval-eccv20} we can also directly optimize the chamfer distance to make our output fit better to the inputs, and we report results with this post-process as well (\textbf{Ours + DO}, \textbf{Ours w/ IDO + DO}).\\

\vspace{-2pt}

\paragraph{Deformation-Aware Retrieval Baseline}
We compare to deformation-aware retrieval~\cite{uy-deformawareretrieval-eccv20} (\textbf{DAR}) followed by either directly optimizing with respect to our per-part parameters (\textbf{DAR+DO}), or using our neural deformation module pre-trained on random pairs (\textbf{DAR+DF}). Note that the direct optimization step is only possible with complete inputs and cannot be employed with partial data such as 3D scans or images. 
Our method outperforms this baseline with and without the direct optimization step (Table~\ref{tbl:pointcloud2cad}). Qualitative results in Figure~\ref{fig:pcicp}, also demonstrate that our method retrieves structurally similar shapes and deforms them to a better fit for the target. Even if retrieved shape is identical (chair in the first row), the deformation learned with our method is superior (e.g., see seat alignment).\\

\paragraph{Template-Classification Baseline}
We also compare to a template-classification-based approach mimicked from~\cite{GanapathiSubramanian2018ParsingGU} (\textbf{Classif}). Instead of using a non-learnable defomation module via direct optimization of handcrafted templates as in~\cite{GanapathiSubramanian2018ParsingGU}, we use our pre-trained neural deformation module (DF) to make the baseline computationally feasible. We treat every source shape as a template, deform it to each training target, and train a classifier based on the best match. We use this classifier instead of the retrieval module at inference time, and show the fitting error in Table~\ref{tbl:pointcloud2cad}. Note that this baseline is worse than our method and even~\cite{uy-deformawareretrieval-eccv20}. \\

\vspace{-2.5pt}


\begin{table}
{
\footnotesize
\centering
\setlength\tabcolsep{2pt}
\begin{tabularx}{\linewidth}{c|C|C|C}
    \toprule
    &Chair&Table&Cabinet\\
    \midrule
    Classif.+DO & 1.826 & 2.192 & 1.144\\
    DAR+DO & 0.584 & 0.452 & 0.633\\
    \textbf{Ours+DO} & \textbf{0.504}  & \textbf{0.414}  & \textbf{0.494} \\
    \textbf{Ours w/ IDO+DO} & \textbf{0.484}  & \textbf{0.407}  & \textbf{0.485} \\
    \midrule
    Classif.+DO & 3.199 & 4.518 & 1.661 \\
    DAR+DF & 0.965 & 1.561  & 0.829 \\
    Uniform Sampling & 0.998  & 1.502 & 0.767 \\
    \textbf{Ours} & \textbf{0.763}  & \textbf{0.696} & 0.\textbf{715} \\
    \textbf{Ours w/ IDO} & \textbf{0.691}  & \textbf{0.670}  & \textbf{0.696} \\
    \bottomrule
\end{tabularx}
\caption{
Comparing our method to various baselines and ablations on points-to-mesh benchmark (chamfer distances, $\times 10^{-2}$).}
\label{tbl:pointcloud2cad}
}
\vspace{-1\baselineskip} 
\end{table}
\paragraph{Biased Sampling Ablation}
As in the image target case, we demonstrate the importance of biased sampling (Equation~\ref{eq:prob}) in joint training (Table~\ref{tbl:pointcloud2cad}, \textbf{Uniform Sampling}).\\
\vspace{-2.5pt}
\begin{figure}[t]
    \centering
    \includegraphics[width=\linewidth]{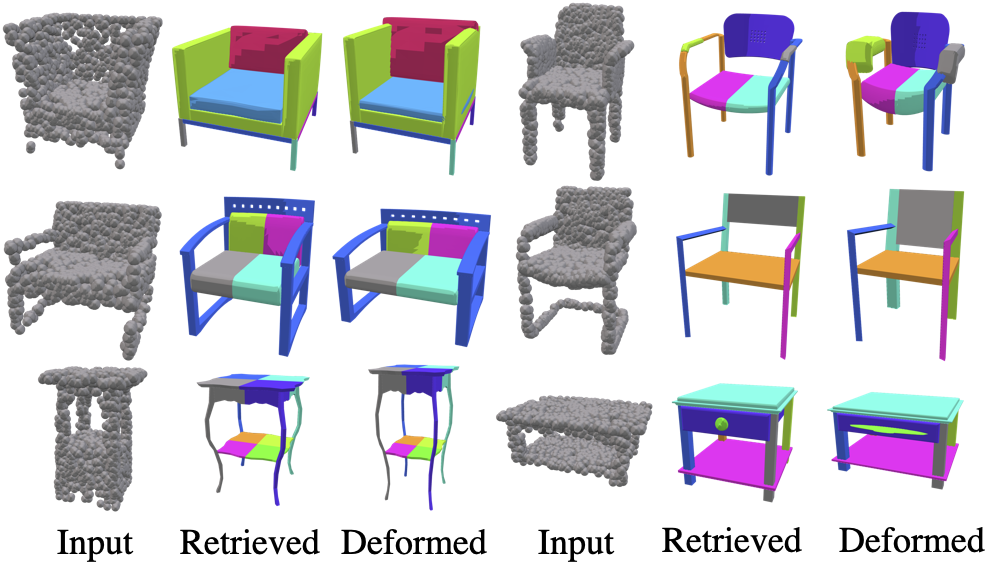}
    \vspace{-6mm} 
    \caption{Fitting results using auto-segmented sources~\cite{Sung:2017}.
    }
    \label{fig:compme}
\end{figure}
\begin{table}
{
\footnotesize
\centering
\setlength\tabcolsep{2pt}
\begin{tabularx}{\linewidth}{C|C|C|C}
    \toprule
    &DAR+DF & {Uniform Sampling} & \textbf{Ours}\\
    \midrule
    Chair & 1.118 & 1.077 & \textbf{0.990} \\
    Table & 1.409 & 1.502 & \textbf{1.166}  \\
    \bottomrule    
\end{tabularx}
\caption{Using auto-segmented models as the source database~\cite{Sung:2017} (chamfer distance ($\times 10^{-2}$).}
\label{tbl:complementme}
}
\end{table}
\paragraph{Performance on Auto-Segmented Data}
Since manually segmenting a collection of source shapes is an expensive process, we test our method on automatically-segmented models.
We use a heuristic method proposed in ComplementMe~\cite{Sung:2017} grouping connected components of meshes.
As shown in Figure~\ref{fig:compme}, even though the models have inconsistent segmentations, our method can still successfully learn a meaningful deformation module. We also outperform the baseline (\textbf{DAR+DF}, \textbf{Uniform Sampling}) in the quantitative benchmark (Table~\ref{tbl:complementme}).\\


\begin{table}
{
\footnotesize
\centering
\setlength\tabcolsep{2pt}
\begin{tabularx}{\linewidth}{c|C|C|C}
    \toprule
    &Chair&Table&Cabinet\\
    \midrule
    DAR+NC & 0.480 & 0.575 & 0.589\\
    Ours NC & \textbf{0.476} & \textbf{0.411} & \textbf{0.538}\\    
    \bottomrule 
\end{tabularx}
\caption{Using our joint training with Neural Cages~\cite{Yifan:NeuralCage:2020} deformation module (chamfer distances, $\times 10^{-2}$).}
\label{tbl:neuralcages}
}
\vspace{-\baselineskip} 
\end{table}

\begin{figure}[t]
    \centering
    \includegraphics[width=\linewidth]{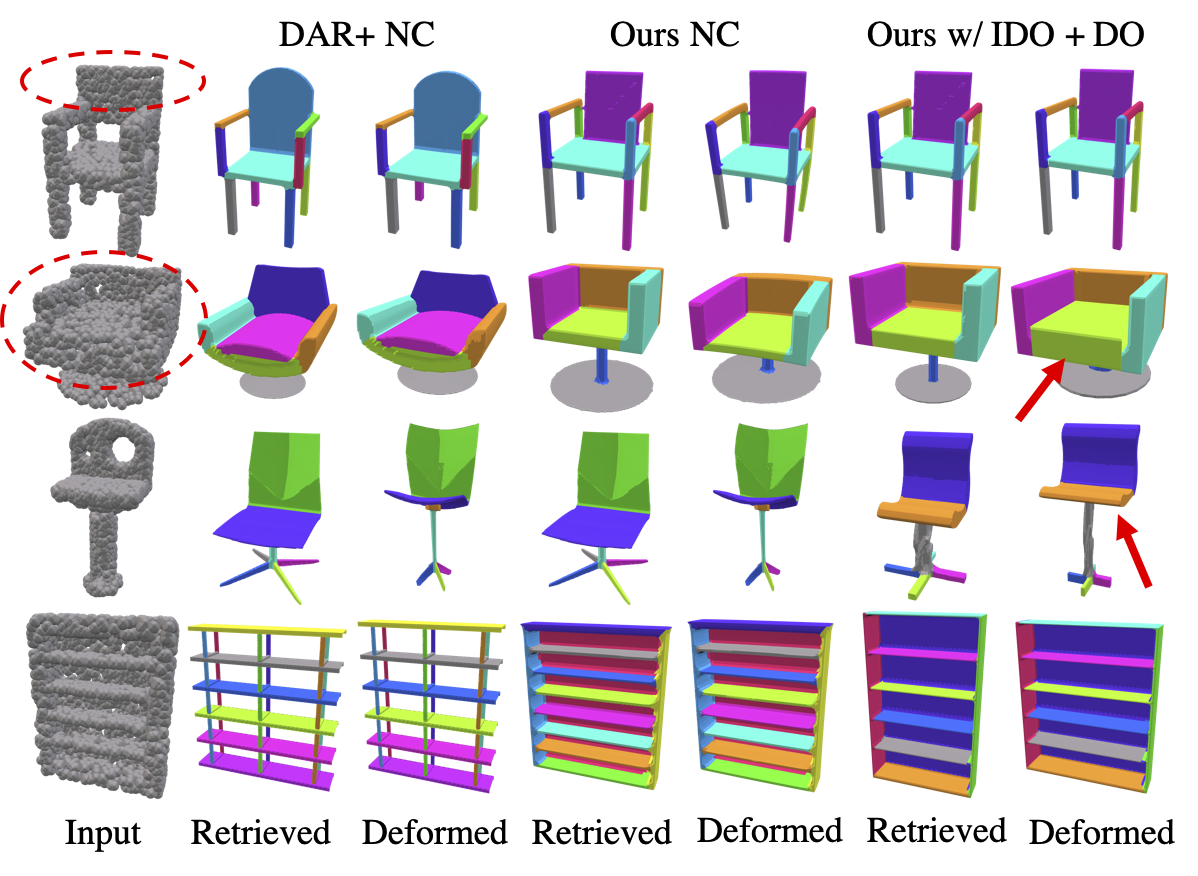}
    \vspace{-6mm} 
    \caption{Using Neural Cages~\cite{Yifan:NeuralCage:2020} as a deformation module in our joint training. 
    }
    \label{fig:nc}
    \vspace{-\baselineskip}
\end{figure}

\vspace{-2pt}
\paragraph{Performance with Neural Cages~\cite{Yifan:NeuralCage:2020}}
\label{nc}
Since our joint training is not restricted to our structure-aware deformation, we further evaluate the performance of our framework with an alternative neural deformation method. We pick Neural Cages~\cite{Yifan:NeuralCage:2020}, a state-of-the-art technique that parameterizes global warping as a cage-based deformation. We simply replace our structure-aware deformation with Neural Cages, without any other changes to our joint training process (\textbf{Ours NC}). We further compare to the baseline of running deformation-aware retrieval~\cite{uy-deformawareretrieval-eccv20} with neural cage module that is pre-trained on random pairs (\textbf{DAR+NC}). Joint training offers an improvement with respect to our benchmark on all categories of shapes (see Table~\ref{tbl:neuralcages}). Qualitative results in Figure~\ref{fig:nc} show that our joint training scheme can better retrieve shapes such as chairs with the right back and seat shape (first two rows), and a cabinet with shelves.

\begin{figure}[t]
    \centering
    \includegraphics[width=\linewidth]{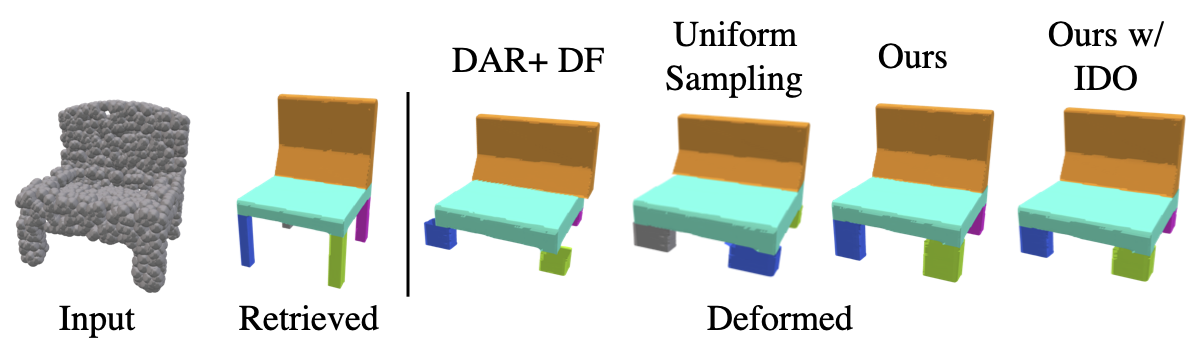}
    \vspace{-6mm} 
    \caption{We pick an example where retrieved mesh is the same for all methods, and show that joint training also improves the quality of the neural deformation module on its own.
    }
    \label{fig:deformcomp}
\end{figure}

We remark that our joint training does not constraint the choice of the neural deformation module. One can choose any module based on its strengths and weaknesses.
For instance, Neural Cages module often provides a tighter fit to the target, although it often results in bending/distortion of shapes (e.g., legs of the chair in the first row and the seat and legs of the chair in the third row of Figure~\ref{fig:nc}). It also lacks the ability to change the geometry of individual local parts. In contrast, our deformation module allows thickening parts such as the seat of the chair in the second row of Figure~\ref{fig:nc}. This implies that Neural Cages can be used when a tighter fit to the target is prioritized while our method can be used when it is more desired to preserve and manipulate part-level structure of the object. Our method is also more suitable for heterogeneous sources whose deformations need to be parameterized in different manners.\\
\vspace{-2pt}
\paragraph{Improvement in Deformation Module}
As in the image target case, we demonstrate the improvement in the deformation module alone using oracle retrieval with joint training (\textbf{Ours}), random pairs (\textbf{DF}), and without biased sampling (\textbf{Uniform Sampling}), see Table~\ref{tbl:deform_point_cloud}. We demonstrate a qualitative example in Figure~\ref{fig:deformcomp} showing an example where all methods retrieve the same source model for the given target, but our joint approach achieves the best output as shown by the differences in the legs of the chair.\\
\vspace{-2pt}

\begin{table}
\footnotesize
\centering
\setlength\tabcolsep{2pt}
\begin{tabularx}{\linewidth}{c|C|C|C}
    \toprule
    & Chair & Table & Cabinet\\
    \midrule
    DF & 0.712 & 0.703 & 0.549 \\
    Uniform Sampling & 0.714  & 0.700  & 0.509 \\
    \textbf{Ours} & \textbf{0.643}  & \textbf{0.564}  & \textbf{0.494} \\
    \textbf{Ours w/ IDO} & \textbf{0.583}  & \textbf{0.482}  & \textbf{0.494} \\
    \bottomrule
\end{tabularx}
\caption{Improvement in deformation module for points-to-mesh (with oracle retrieval) due to joint training (chamfer distances, $\times 10^{-2}$).
}
\label{tbl:deform_point_cloud}
\vspace{-\baselineskip} 
\end{table}

\begin{table}
{
\footnotesize
\centering
\begin{tabularx}{\linewidth}{L|C|C|C}
    \toprule
    & DAR+DF &Uniform Sampling&\textbf{Ours}\\
    \midrule
    $|\mathbf{S}|$=50 & 0.872  & 0.877  & \textbf{0.823} \\
    $|\mathbf{S}|$=100 & 0.858 & 0.860 & \textbf{0.803} \\
    $|\mathbf{S}|$=200 & 0.850 & 0.841 & \textbf{0.748}\\
    $|\mathbf{S}|$=400  & 0.938 & 0.985 & \textbf{0.784} \\
    $|\mathbf{S}|$=800 & 1.142 & 1.541 & \textbf{0.734} \\
    \bottomrule    
\end{tabularx}
\caption{Performance on of our method and various baselines with different source database sizes (chamfer distances, $\times 10^{-2}$).}
\label{tbl:diff_data_sizes}
}
\vspace{-\baselineskip} 
\end{table}

\paragraph{Performance for Different Database Sizes}
We further evaluate the performance of different techniques while varying the size of the database of source models. We randomly sample 50, 100, 200, 400, and 800 chair models from PartNet to construct the source databases. Table~\ref{tbl:diff_data_sizes} shows that in all cases our joint training approach improves the performance over the baselines. The boost in the performance of our joint training is bigger in larger databases as there are combinatorially more random source-target pairs which may not be deformable.
\section{Conclusion}
\vspace{-3pt}
To summarize, we propose a joint training for retrieval-and-deformation problem, where the neural modules inform one another, yielding better matching results with respect to image and point cloud targets. Our joint training procedure offers improvements regardless of the choice of the neural deformation module. We further propose a novel structure-aware deformation module that is especially suitable for hetereogeneous datasets of source models with very diverse parameterizations of deformations. Our method does not require consistent manual segmentations or part labels and can work with imprecise automatic segmentations. 

\paragraph{Limitations and Future Work}
Our method is only supervised by chamfer distance, and thus might not favor semantic and structural similarity between the target and retrieved sources. We believe that improving the loss function to leverage manual part annotations can further remedy this issue. Our deformation module does not provide strong links between parts, and does not favor capturing part-to-part relations, which can be addressed by adding more constraints (\eg symmetry) as well as improving our learning module with a more advanced graph-based neural architecture. 

\paragraph{Acknowledgements} This work is supported by a grant from the Samsung GRO program, a Vannevar Bush Faculty Fellowship, and gifts from Adobe, Autodesk, and Snap.

{\small
\bibliographystyle{ieee_fullname}
\bibliography{egbib}
}

\clearpage

\renewcommand{\thesection}{A}
\setcounter{table}{0}
\renewcommand{\thetable}{A\arabic{table}}
\setcounter{figure}{0}
\renewcommand{\thefigure}{A\arabic{figure}}

\newif\ifpaper
\papertrue

\section*{Appendix}
\ifpaper
  \newcommand\refpaper[1]{\unskip}
\else
  \makeatletter
  \newcommand{\manuallabel}[2]{\def\@currentlabel{#2}\label{#1}}
  \makeatother
  \manuallabel{fig:results}{4}
  \manuallabel{eq:loss_embed}{4}
  \manuallabel{sec:joint}{3.1}
  \manuallabel{sec:deform}{3.2}
  \manuallabel{tbl:image2cad}{1}
  \manuallabel{tbl:pointcloud2cad}{3}
  \manuallabel{tbl:deform_image}{2}
  \manuallabel{tbl:deform_point_cloud}{6}
  \newcommand{\refpaper}[1]{in the paper}
\fi

We provide additional implementation details  (Section~\ref{sec:method_details}), and additional quantitative evaluations (Section~\ref{sec:add_quantitative}) and qualitative results (Section~\ref{sec:add_qualitative}).

\subsection{Implementation Details}
\label{sec:method_details}
\noindent \textbf{Inner Deformation Optimization.}
We provide additional details for the inner deformation optimization step, as described in Section~\ref{sec:joint} of the main paper. 

We initialize the inner deformation optimization with the parameters predicted by our deformation network. We propagate gradients directly to the parameters by minimizing the mean chamfer loss of the batch. We use the SGD optimizer with a learning rate of 0.05, and we terminate upon convergence (i.e., when the maximum loss change in a pair in the batch is less than $10^{-6}$ or it has reach the maximum number of iterations $=2000$). \\

\noindent \textbf{Structure-Aware Neural Deformation.}
We provide additional details for our structure-aware neural deformation as described in Section~\ref{sec:deform} of the main paper. 

Our structure-aware neural deformation module predicts the deformation parameter offset from the default parameters of each source model. Specifically for a specific source-target pair, given network prediction $p$ and default source parameter $\bar{p}$, our output parameters to obtain the deformed source model is given by $(\bar{p}+\alpha*p$, where $\alpha=0.1)$ in all our experiments. 

We also add the symmetry loss to supervise the training of our structure-aware neural deformation. Note that all the source shapes in our databases have global reflective symmetry, and have been pre-aligned so that yz-plane aligns with the symmetry axis. Given the output deformed source shape, represented as a sampled point cloud $O$, for target point cloud $T$ of given target $\mathbf{t}$, we reflect each point $O$ about the yz-plane to obtain reflected point cloud $O'$, then the symmetry loss is given by
$$\mathcal{L}_{\text{symm}}=\mathcal{L}_{\text{CD}}(O, O'),$$
\noindent where $\mathcal{L}_{\text{CD}}$ is the chamfer distance. Then the loss we use to train our deformation module is given by
$$\mathcal{L}_{\text{total}} = \mathcal{L}_{\text{def}}  + \mathcal{L}_{\text{symm}},$$
\noindent where $\mathcal{L}_{\text{def}}$ is defined in Equation~\ref{eq:loss_embed} in the main paper. \\

\noindent \textbf{Connectivity constraint.}
We provide the details on how we obtain our connectivity constraint as described in Section~\ref{sec:deform} of the main paper. 

We precompute the constraint projection matrix for each source $\mathbf{s} \in \mathbf{S}$ in an automatic pre-processing step, where we first identify contacts based on the distance between the closest pairs of keypoints between pairs of parts $(\mathbf{s}_\mathcal{D}^i, \mathbf{s}_\mathcal{D}^j)$. Parts $\mathbf{s}_\mathcal{D}^i$ and $\mathbf{s}_\mathcal{D}^j$ are deemed connected if the closest part of keypoints falls below a threshold $\tau=0.05$. Part keypoints is the set of face centers, edge midpoints, and corners of each part's axis-aligned bounding box. We then define contacts as the midpoint of the closest pair of keypoints of two connected parts, and obtain 3 linear constraints (one for each axis) for each pair of connected parts that enforces the contact point to maintain connectivity during deformation. We obtain a number of linear constraints from the collection of contacts that results in a different number of linear constraints for each source model. We concatenate all the linear constraints and represent these with constraint matrix $B_\mathbf{s}$ for source model $\mathbf{s}$. Let $Q_\mathbf{s}$ be the nullspace, \ie columns representing the nullspace basis vectors, of $B_\mathbf{s}$ computed via SVD, then the constraint projection matrix of $\mathbf{s}$ is given by $Q_\mathbf{s}Q_\mathbf{s}^T$. \\

\noindent \textbf{Training details and training time.}
We alternately update the retrieval module and the deformation module at each iteration during our training procedure, and train for 300 epochs. To speedup training, we cache the distances to the sources for each target and update this cache every 5 epochs. We use a batch size of 16 targets in each iteration, the SGD optimizer with learning rate of 0.001, momentum of 0.9 and weight decay of 0.0005. For the inner deformation optimization, also use the SGD optimizer with a learning rate of 0.05 until the termination criteria is reached, which is when the fitting loss decreases by less than $10^{-5}$ or the maximum number of 5000 iterations is reached.

For our joint training module, we first train our Structure-Aware neural deformation module until convergence on random pairs, and also train our retrieval module on random pairs to initialize our joint training optimization scheme. Also note that when training image-based ResNet encoder for the retrieval and deformation modules, we warm-start with weights that are pre-trained on ImageNet, and only train the fourth block and the final fully-connected layers.

Training takes $~18$  and  $~40$  hours   on  point  clouds and  images, respectively, for the chair class. With the  inner  loop  direct  optimization, the  corresponding  training  time  for  chairs  takes $~3$ days for both the point cloud and image experiments as the inner optimization dominates the runtime.

\subsection{Additional Results}
\subsubsection{Additional Quantitative Evaluations}
\label{sec:add_quantitative}
\begin{figure*}[t]
    \centering
    \includegraphics[width=\linewidth]{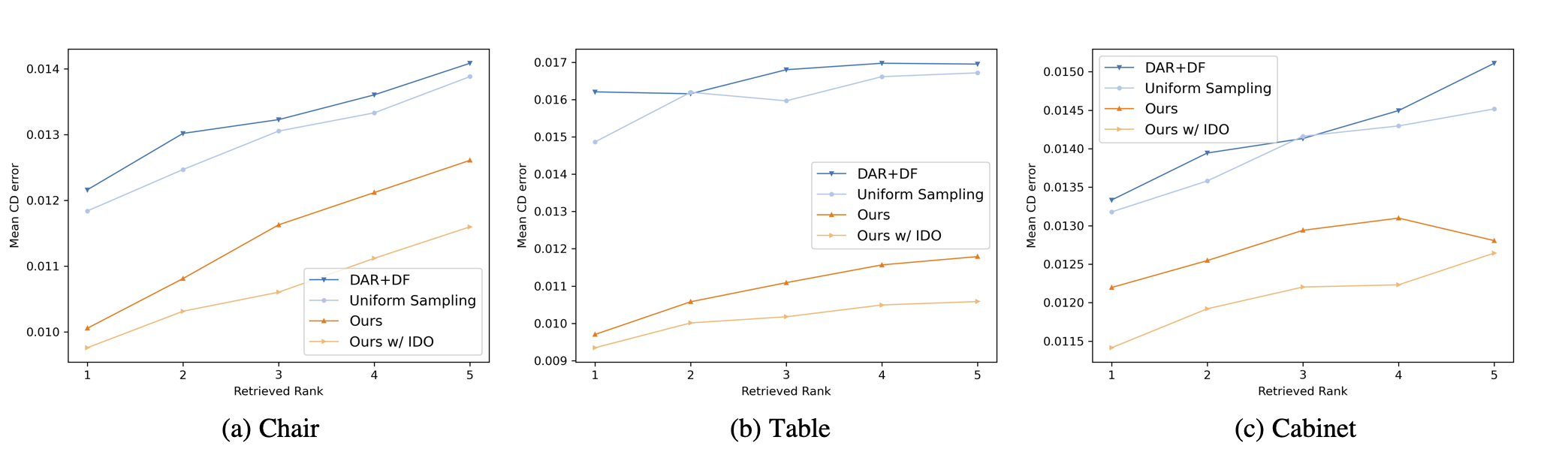}
    \caption{Quantitative evaluation of Image-to-Mesh.}
    \label{fig:supp_quant_image}
    \vspace{-\baselineskip}
\end{figure*}

\begin{figure*}[t]
    \centering
    \includegraphics[width=\linewidth]{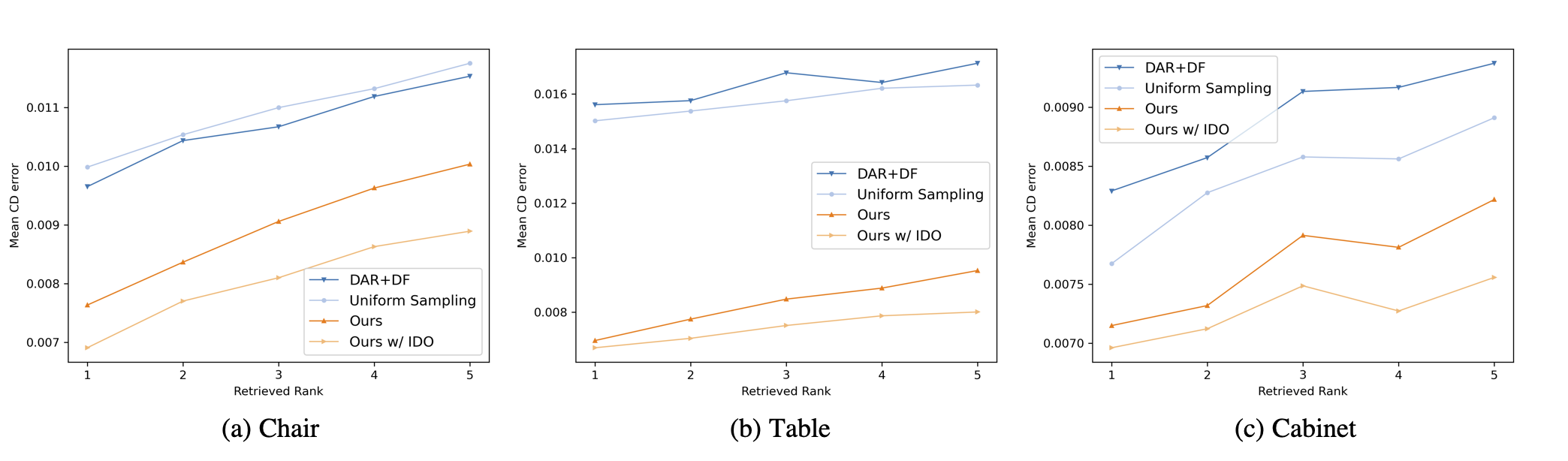}
    \caption{Quantitative evaluation of Points-to-Mesh.}
    \label{fig:supp_quant_pc}
\end{figure*}

\begin{table}
{
\footnotesize
\centering
\setlength\tabcolsep{2pt}
\begin{tabularx}{\linewidth}{c|C|C|C}
    \toprule
    & Chair & Table & Cabinet \\
    \midrule
    DAR+DF (No Conn.) & 1.107 & 1.728 & 1.480 \\
    Uniform Sampling (No Conn.) & 1.129  & 1.655  & 1.358 \\
    \textbf{Ours} (No Conn.) & \textbf{0.757}  & \textbf{0.708}  & \textbf{0.846} \\
    \bottomrule 
\end{tabularx}
\caption{Our approach compared to the baselines in the setup with no connectivity constraint. }
\label{tbl:no_conn}
}
\vspace{-\baselineskip} 
\end{table}

\noindent \textbf{No connectivity constraint ablation.}
We also test our joint training scheme in the setting where the source database models do not have connectivity constraints. In this set-up we do not use the constraint projection matrix. Table~\ref{tbl:no_conn} shows that even in the set-up with no connectivity, our approach achieves the best results in all three object classes. \\

\noindent \textbf{Retrieval-and-deformation results for different retrieved sources.}
We further evaluate how well our method works with other than top-1 retrieved source. In particular, we plot the mean chamfer distance for the $k^\text{th}$ retrieved source, for $k=1,2,3,4,5$. 

For image-to-mesh experiment, we show the result in Figure~\ref{fig:supp_quant_image}, which complements Table~\ref{tbl:image2cad} of the main paper. For points-to-mesh experiment, we show the result in Figure~\ref{fig:supp_quant_pc}, which complements Table~\ref{tbl:pointcloud2cad} of the main paper. Note that in both cases the chamfer distance for up to top-5 retrieved results is consistently lower than the baselines. \\

\noindent \textbf{Retrieval module evaluation.}
We further evaluate the retrieval modules of our joint approach compared to the baselines. To evaluate the retrieval module, we report both \emph{ranking evaluation} and \emph{recall} similar to the metrics used in ~\cite{uy-deformawareretrieval-eccv20}.  

One challenge in defining an evaluation metric is that we do not know which source model should be used for each target. Thus, to create the ground truth we use \emph{oracle retrieval}, where we use the each method's deformation module to deform each source to the target, and assume that if we sort the sources by the chamfer distance, it will give us the desired ground truth ordering for the retrieval. 

Ranking evaluation reports the average rank of the top-1 retrieved model with respect to this ground truth. We report the metrics for image-to-mesh (Table~\ref{tbl:image_ranking}) and points-to-mesh (Table~\ref{tbl:points_ranking}) experiments, across all categories, and see consistent improvement with respect to the baselines. 

We also report the recall of retrieval modules. For recall@$N$, a correct match is defined as the case where at least one of the top-$N$ retrieved models is in the top-5 ranks based on the oracle retrieval module. We report both recall@1 and recall@5. We report the metrics for image-to-mesh (Table~\ref{tbl:image_recall}) and points-to-mesh (Table~\ref{tbl:points_recall}) experiments, across all categories, and see consistent improvement with respect to the baselines.






\vspace{0.4cm}
\noindent \textbf{Additional object categories.} We ran experiments on additional categories (vases, beds, trash cans), and a combination of categories (chairs+tables+cabinets). As shown in Table~\ref{tbl:obj_class}, we got a comparable performance and improvement over baselines.

\begin{table}
{
\footnotesize
\centering
\setlength\tabcolsep{2pt}
\begin{tabularx}{\linewidth}{c|C|C|C}
    \toprule
    &Chair&Table&Cabinet\\
    \midrule
    DAR+DF & 23.98 & 59.51  & 19.50 \\
    Uniform Sampling & 20.88 & 53.01 & 23.39 \\
    \textbf{Ours} & \textbf{15.35}  & \textbf{22.19} & 21.70 \\
    \textbf{Ours w/ IDO} & 21.94  &  \textbf{36.92} & \textbf{16.89} \\
    \bottomrule
\end{tabularx}
\caption{
\textbf{Ranking evaluation for retrieval.} Comparing our method using the ranking evaluation metric on image-to-mesh benchmark. Numbers show the average rank of the retrieved model. (Lower is better)}
\label{tbl:image_ranking}
}
\end{table}

\begin{table}
{
\footnotesize
\centering
\setlength\tabcolsep{2pt}
\begin{tabularx}{\linewidth}{c|C|C|C}
    \toprule
    &Chair&Table&Cabinet\\
    \midrule
    DAR+DF & 13.88 & 76.25  & 20.20 \\
    Uniform Sampling & 18.27 & 72.44 & 23.44 \\
    \textbf{Ours} & \textbf{6.37}  & \textbf{6.97} & \textbf{17.91} \\
    \textbf{Ours w/ IDO} & \textbf{6.62}  & \textbf{18.03}  & \textbf{18.22} \\
    \bottomrule
\end{tabularx}
\caption{
\textbf{Ranking evaluation for retrieval.} Comparing our method using the ranking evaluation metric on points-to-mesh benchmark. Numbers show the average rank of the retrieved model. (Lower is better)}
\label{tbl:points_ranking}
}
\end{table}

\begin{table}
{
\footnotesize
\centering
\setlength\tabcolsep{2pt}
\begin{tabularx}{\linewidth}{c|CC|CC|CC}
    \toprule
    &\multicolumn{2}{c|}{Chair}&\multicolumn{2}{c|}{Table}&\multicolumn{2}{c}{Cabinet}\\
    \cline{2-7}
    & \scriptsize{recall@1} & \scriptsize{recall@5}  & \scriptsize{recall@1} & \scriptsize{recall@5}  & \scriptsize{recall@1} & \scriptsize{recall@5}  \\
    \midrule    
    \midrule
    DAR+DF & 37.53 & 74.65 & 14.55 & 43.46 & 22.37 & 57.89\\
    Uniform Sampling & 38.94 & 75.56 & 21.90 & 54.79 & 21.05 & 53.81\\
    \textbf{Ours} & \textbf{53.60} & \textbf{81.03} & \textbf{53.81} & \textbf{82.93} & \textbf{30.70} & \textbf{61.40} \\
    \textbf{Ours w/ IDO} & \textbf{45.65} & \textbf{77.30} & \textbf{35.83} & \textbf{69.35} &\textbf{ 35.96} &\textbf{65.79} \\
    \bottomrule
\end{tabularx}
\caption{
\textbf{Recall evaluation for retrieval.} Comparing our method using the ranking evaluation metric on image-to-mesh benchmark. Numbers show recall@1 and recall@5. A correct retrieval is when the top-1 and top-5 retrieved models is in the top-5 ranks based on the oracle retrieval. (Higher is better)}
\label{tbl:image_recall}
}
\end{table}

\begin{table}
{
\footnotesize
\centering
\setlength\tabcolsep{2pt}
\begin{tabularx}{\linewidth}{c|CC|CC|CC}
    \toprule
    &\multicolumn{2}{c|}{Chair}&\multicolumn{2}{c|}{Table}&\multicolumn{2}{c}{Cabinet}\\
    \cline{2-7}
    & \scriptsize{recall@1} & \scriptsize{recall@5}  & \scriptsize{recall@1} & \scriptsize{recall@5}  & \scriptsize{recall@1} & \scriptsize{recall@5}  \\
    \midrule    
    \midrule
    DAR+DF & 61.56 & 93.54 & 23.57 & 54.54 & 39.83 & 72.29 \\
    Uniform Sampling & 53.27 & 89.98 & 25.03 & 59.16 & 39.83 & 67.97 \\
    \textbf{Ours} & \textbf{75.31} & \textbf{97.02} & \textbf{73.71} & \textbf{96.50} & \textbf{48.05} & \textbf{76.19} \\
    \textbf{Ours w/ IDO} & \textbf{76.22} & \textbf{96.60} & \textbf{55.17} & \textbf{89.72} & 38.53 & \textbf{77.06} \\
    \bottomrule
\end{tabularx}
\caption{
\textbf{Recall evaluation for retrieval.} Comparing our method using the ranking evaluation metric on points-to-mesh benchmark. Numbers show recall@1 and recall@5. A correct retrieval is when the top-1 and top-5 retrieved models is in the top-5 ranks based on the oracle retrieval. (Higher is better)}
\label{tbl:points_recall}
}
\end{table}

\begin{table}
\footnotesize
\setlength\tabcolsep{2pt}
\begin{tabularx}{\linewidth}{c|CCCC}
	\toprule
	&Vase&Bed&Trash Can&Combined\\
	\midrule
	DAR+DF & 1.538 & 4.498 &  0.889 & 1.968\\
	Uniform Sampling & 1.633  & 4.196 & 0.886 & 1.821\\
	\textbf{Ours} & \textbf{1.384}  & \textbf{2.138} & \textbf{0.863} & \textbf{0.810}\\
	\bottomrule
\end{tabularx}
\caption{\textbf{Additional object categories.} Comparing our method to various baselines and ablations on additional object classes and mixture of categories (chamfer distances, $\times 10^{-2}$).}
\label{tbl:obj_class}

\vspace{-1.5\baselineskip} 
\end{table}

\begin{figure}[t]
    \centering
    \includegraphics[width=0.9\linewidth]{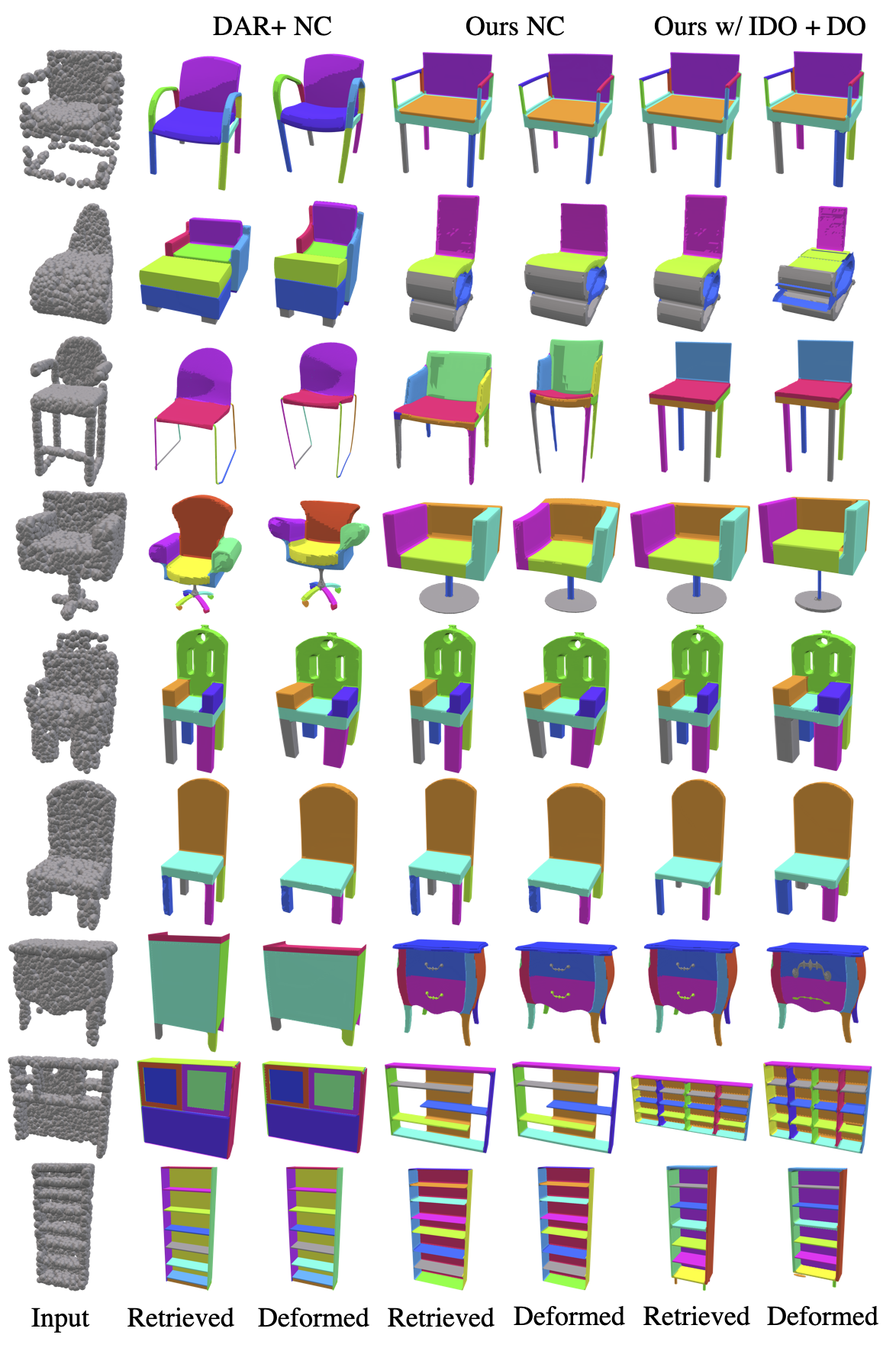}
    \caption{More qualitative results on Neural Cages~\cite{Yifan:NeuralCage:2020}.
    }
    \label{fig:supp_neural cages}
\end{figure}

\begin{figure}[t]
    \centering
    \includegraphics[width=0.9\linewidth]{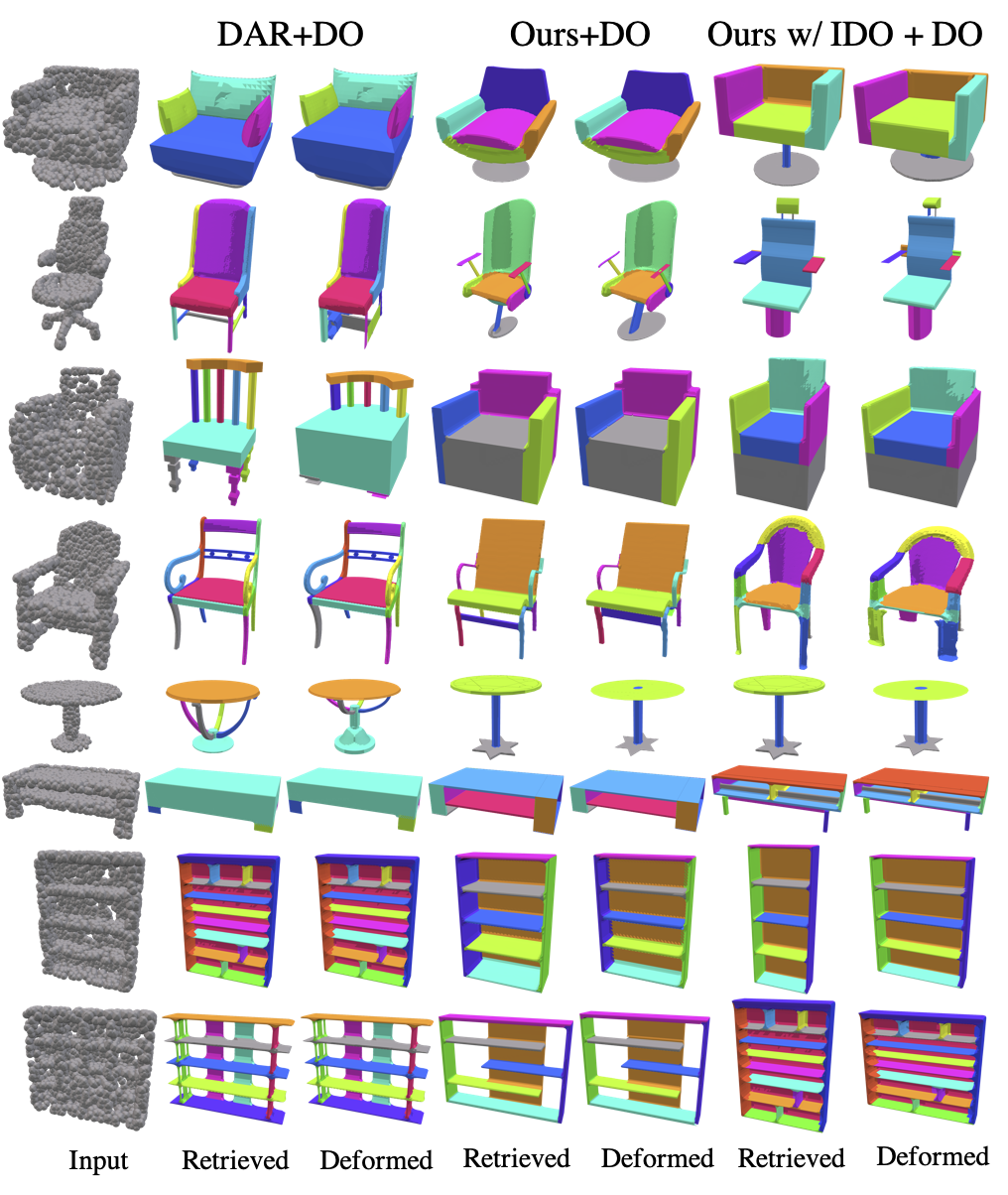}
    \caption{Additional qualitative results on comparisons between our approach and the baselines for the points-to-mesh experiments.
    }
    \label{fig:pc-icp}
\end{figure}

\vspace{0.4cm}
\noindent \textbf{Perceptual Metric.} We performed a user study comparing our approach to the DAR+DF baseline. We asked 60 participants to pick the better match to input point clouds on 15 randomly selected targets from the test set, where an option of ``no difference" can also be selected. Our approach got an average score of \textbf{8.02}, compared to 3.5 for the baseline and 3.48 abstain votes. 

\subsubsection{Additional Qualitative Results}
\label{sec:add_qualitative}
We provide additional qualitative results using natural images, point cloud scans, and our benchmark as input targets. Note that in all visualizations, we use colors to indicate different segmentations of the source models, where segmentation is essential to the performance of the structure-aware neural deformation module.  \\
%

\begin{figure*}[t]
    \centering
    \includegraphics[width=\linewidth]{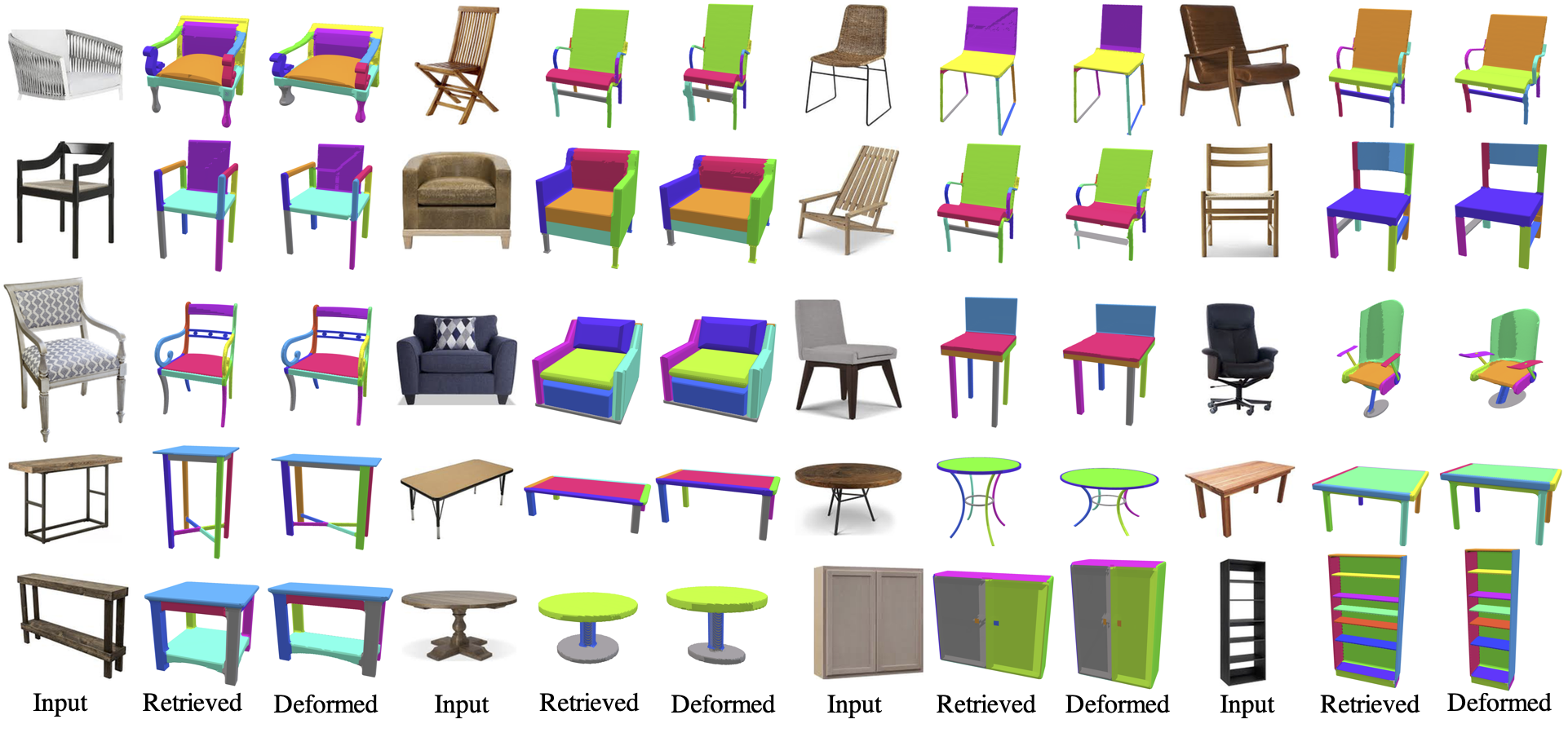}
    \caption{More qualitative results on product images.
    }
    \label{fig:supp_natural_images}
    \vspace{-\baselineskip}
\end{figure*}

\noindent \textbf{Product images targets.}
Figure~\ref{fig:supp_natural_images} shows additional qualitative results of our approach on product images. \\

\begin{figure*}[t]
    \centering
    \includegraphics[width=\linewidth]{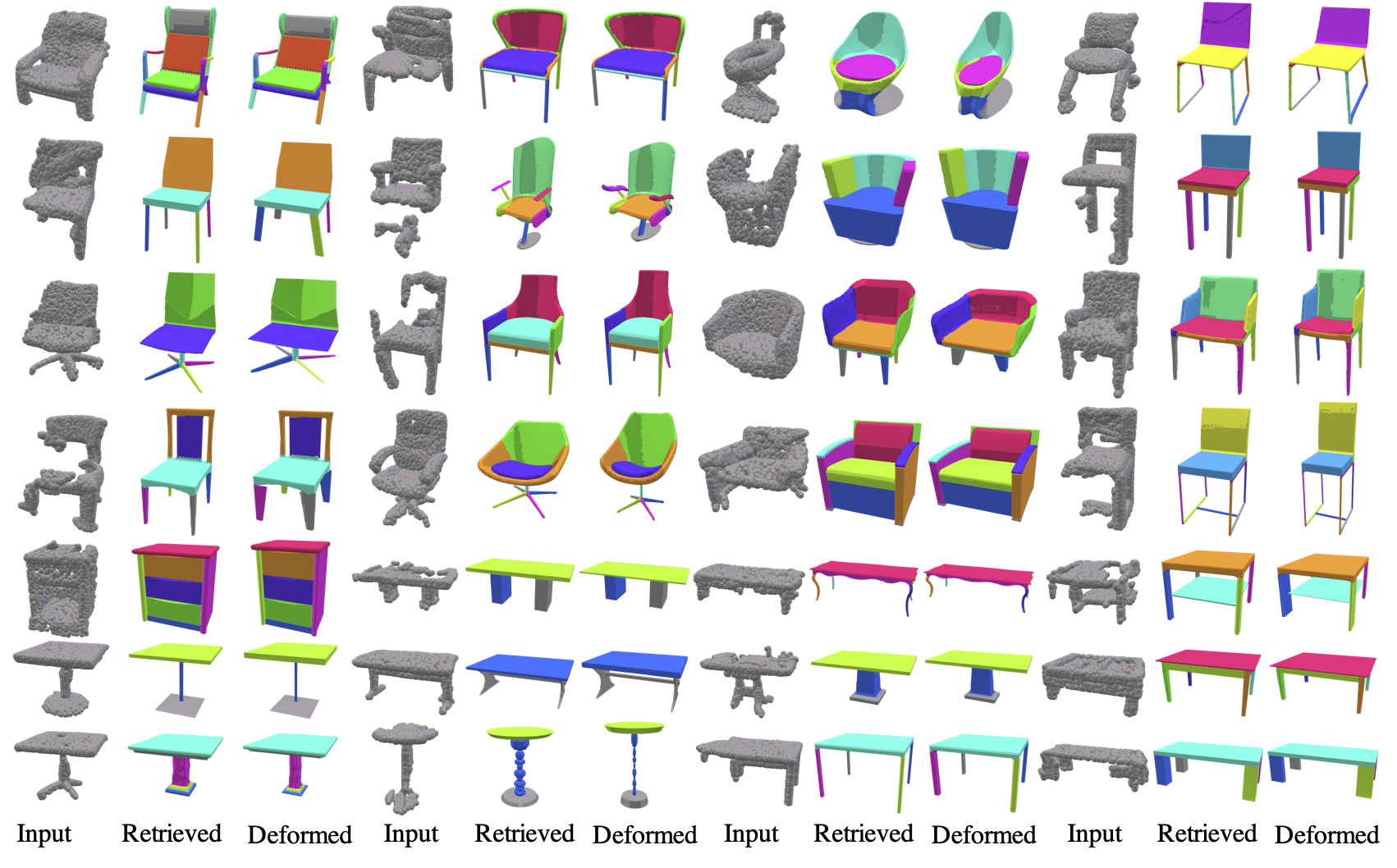}
    \caption{More qualitative results using the Scan2CAD~\cite{Avetisyan:2019:Scan2CAD} dataset using manually segmented shapes in PartNet~\cite{Mo_2019_CVPR}.
    }
    \label{fig:supp_scan2cad}
    \vspace{-\baselineskip}
\end{figure*}

\begin{figure*}[t]
    \centering
    \includegraphics[width=\linewidth]{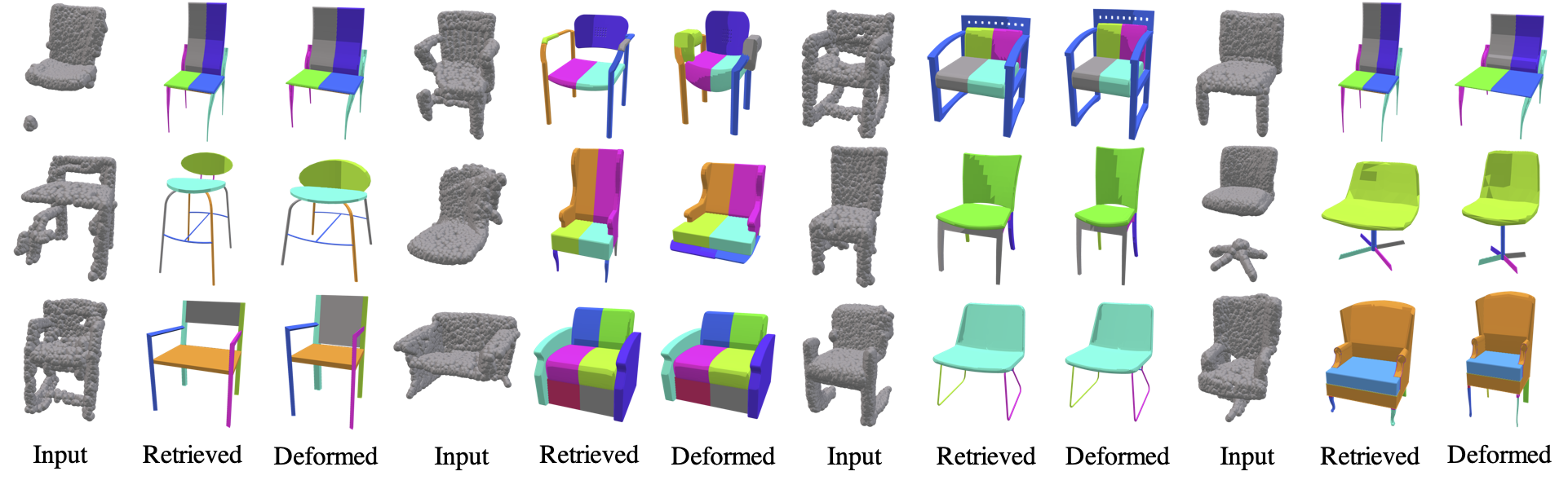}
    \caption{More qualitative results using the Scan2CAD~\cite{Avetisyan:2019:Scan2CAD} dataset using autosegmented shapes in ComplementMe~\cite{Sung:2017}.
    }
    \label{fig:supp_scan2cadcompme}
    \vspace{-\baselineskip}
\end{figure*}

\noindent \textbf{Scan2CAD targets.}
Figure~\ref{fig:supp_scan2cad} shows additional results of our approach on real scans from the Scan2CAD~\cite{Avetisyan:2019:Scan2CAD} dataset using the manually segmented PartNet~\cite{Mo_2019_CVPR} database, while Figure~\ref{fig:supp_scan2cadcompme} shows the results on real scans using the auto-segmented ComplementMe~\cite{Sung:2017} database. \\

\begin{figure*}[t]
    \centering
    \includegraphics[width=0.77\linewidth]{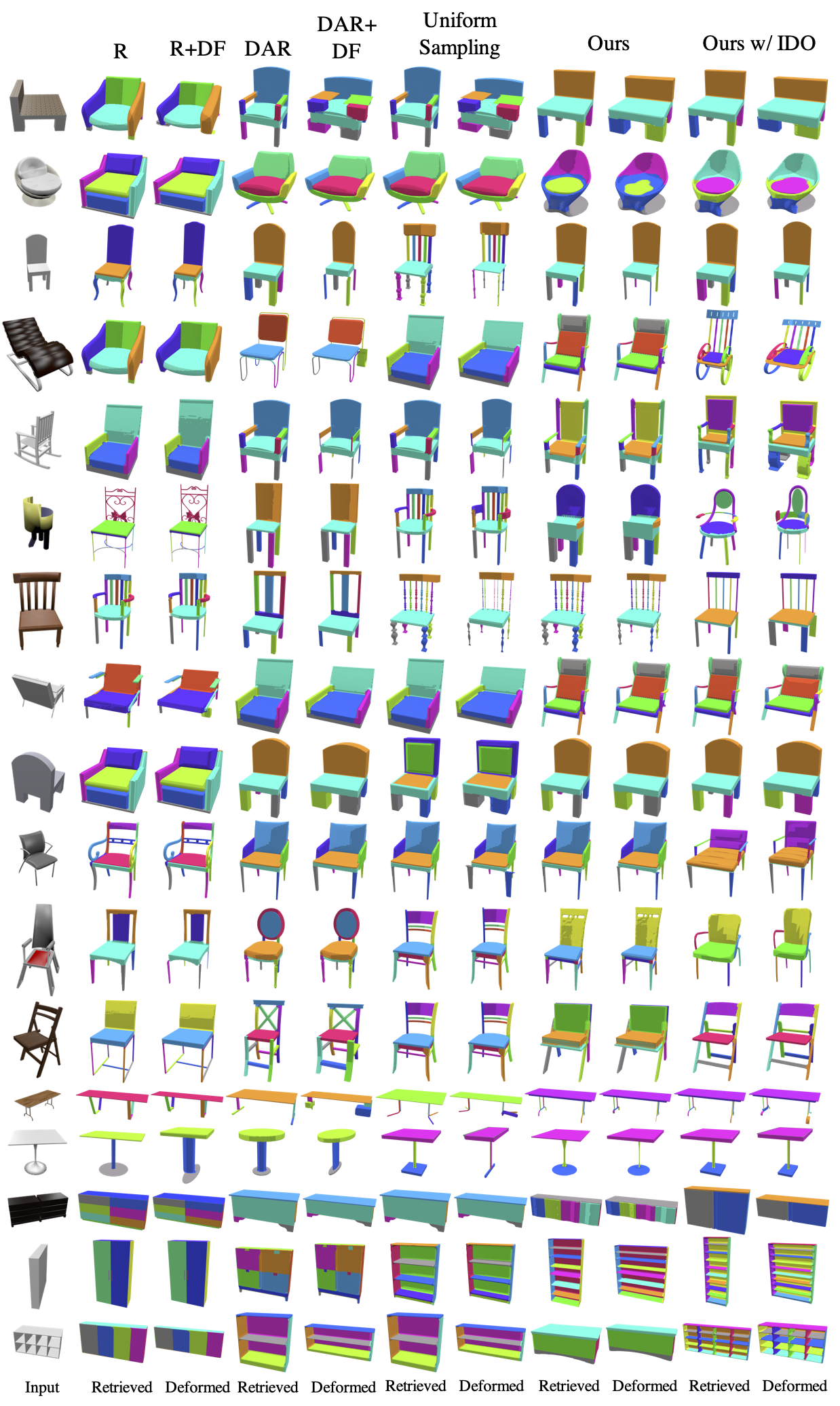}
    \caption{More qualitative results on Image-to-Mesh.
    }
    \label{fig:supp_images}
    \vspace{-\baselineskip}
\end{figure*}

\noindent \textbf{Image-to-Mesh baseline comparison.}
Figure~\ref{fig:supp_images} shows additional qualitative results on the image-to-mesh set-up that compares our method to the baselines. \\

\noindent \textbf{Points-to-Mesh baseline comparison.}
Figure~\ref{fig:pc-icp} shows additional qualitative results of our joint approach compared to the baselines on the points-to-mesh experiment. \\

\noindent \textbf{Neural cages.}
Figure~\ref{fig:supp_neural cages} shows additional qualitative results of our joint approach on Neural Cages~\cite{Yifan:NeuralCage:2020}. \\

\begin{figure*}[t]
    \centering
    \includegraphics[width=0.85\linewidth]{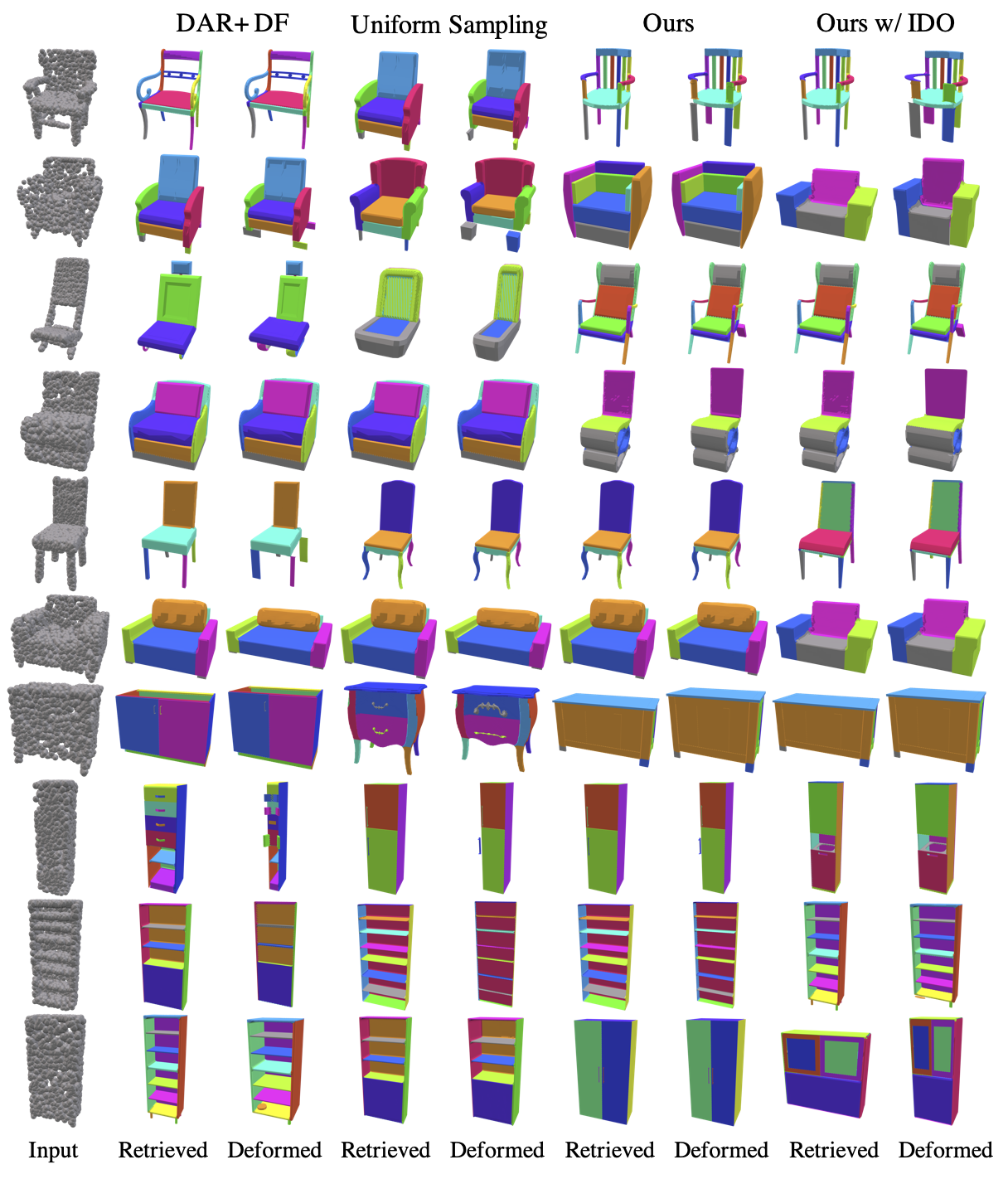}
    \caption{More qualitative results on Points-to-Mesh.
    }
    \label{fig:supp_points}
    \vspace{-\baselineskip}
\end{figure*}

\noindent \textbf{Points-to-Mesh ablations qualitative results.}
Figure~\ref{fig:supp_points} shows qualitative results of ablations of our joint approach on the points-to-mesh experiment. \\

\subsection{Discussion on~\cite{GanapathiSubramanian2018ParsingGU}}

\noindent The differences between our work and with~\cite{GanapathiSubramanian2018ParsingGU} are as follows:

\vspace{-0.5\baselineskip}
\begin{enumerate}[leftmargin=*]
	\setlength{\itemsep}{0pt}
	\setlength{\parskip}{0pt}
	
	\item \textbf{Non-learnable deformations}: The fitting module of~\cite{GanapathiSubramanian2018ParsingGU} is \emph{not learnable}; they directly \emph{optimize} parameters of a handcrafted template to fit to an input point cloud. Thus, one of our key contributions, a \emph{retrieval-aware deformation}, is incompatible with their method.  
	
	\item \textbf{Infeasibility of image-to-mesh}: Without learnable deformations, their method cannot be used for the main application of our method, image-to-mesh generation.
	
	\item \textbf{Manually-designed templates}: Designing templates is a tedious manual task that requires significant expertise. Their method requires users to pre-design a set of templates, hence they only use a small set of 21 templates.
	
	\item \textbf{Non-scalable system}: While one could address solving our retrieval problem as a classification problem by treating every source shape as a template, this approach is not scalable. Their method requires a pre-process of matching every template to every input shape for training. Their optimization-based deformation module takes 2-3 mins for a single pair, and thus for all 500 sources and 4000 training targets as in our chair dataset, it would take $\sim 8$ years. Note that this limitation has been addressed in a recent work of Uy et al.~\cite{uy-deformawareretrieval-eccv20} who propose to learn a \emph{deformation-aware retrieval} latent space instead of the non-scalable hard shape-to-template assignment (and we extensively compared to Uy et al.~\cite{uy-deformawareretrieval-eccv20}).
	
	\item \textbf{Specific to template-based deformations}: Our key contribution, \emph{joint} learning for retrieval and deformation, is not constrained to a specific choice of the deformation module.
	
\end{enumerate}
We also remark that, while both ours and their method leverage on part bounding boxes for deformations, neither of these two were the first to use bounding boxes to deform the underlying geometry (e.g., ~\cite{Kim13}). \\

\end{document}